%% file: main.tex
\documentclass[runningheads]{llncs}

\input{math_commands.tex}

\usepackage[utf8]{inputenc} 
\usepackage[T1]{fontenc}    
\usepackage{url}            
\usepackage{booktabs}       
\usepackage{amsfonts}       
\usepackage{nicefrac}       
\usepackage{microtype}      
\usepackage{graphicx}
\usepackage{amsmath,amssymb}
\usepackage{multirow}
\usepackage{makecell}
\usepackage{url}
\usepackage{wrapfig}

\usepackage{tikz}
\usepackage{comment}
\usepackage{color}
\usepackage[title]{appendix}

\usepackage[accsupp]{axessibility}  

\usepackage[utf8]{inputenc}
\usepackage[english]{babel}
\usepackage{tablefootnote}
\usepackage{algorithmic}
\usepackage[linesnumbered,ruled,vlined]{algorithm2e}

\definecolor{myblue}{RGB}{0, 128, 255}
\definecolor{mycolor}{RGB}{147,112,219}

\definecolor{citecolor}{HTML}{0071bc}

\newcommand{\yc}[1]{{\color{myblue}{(YC: #1)}}}
\newcommand{\zk}[1]{{\color{red}{(ZK: #1)}}}
\newcommand{\kevin}[1]{{\color{mycolor}{(Kevin: #1)}}}
\newcommand{\kevinedit}[1]{{\color{citecolor}{(Kevin: #1)}}}
\newcommand{\jtian}[1]{{\color{blue}{(jtian: #1)}}}

\renewcommand{\yc}[1]{{}}
\renewcommand{\zk}[1]{{}}
\renewcommand{\kevin}[1]{{}}
\renewcommand{\kevinedit}[1]{{}}
\renewcommand{\jtian}[1]{{}}

\newcommand{\opensetissue}{\textit{semantic expansion}}

\usepackage[pagebackref,breaklinks,colorlinks, citecolor=citecolor]{hyperref}

\usepackage[capitalize]{cleveref}
\crefname{section}{Sec.}{Secs.}
\Crefname{section}{Section}{Sections}
\Crefname{table}{Table}{Tables}
\crefname{table}{Tab.}{Tabs.}

\begin{document}

\pagestyle{headings}
\mainmatter
\def\ECCVSubNumber{617}  

\title{Open-Set Semi-Supervised Object Detection}

%

\titlerunning{Open-Set Semi-Supervised Object Detection}
%
\author{Yen-Cheng Liu\inst{1}\thanks{Work done partially while interning at Meta.} \and
Chih-Yao Ma\inst{2} \and
Xiaoliang Dai\inst{2} \and
Junjiao Tian\inst{1} \and \\
Peter Vajda\inst{2} \and
Zijian He\inst{2} \and
Zsolt Kira\inst{1}
}
\authorrunning{Liu et al.}
%
\institute{$^{1}$Georgia Tech, $^{2}$Meta}
\maketitle


\begin{abstract}
Recent developments for Semi-Supervised Object Detection (SSOD) have shown the promise of leveraging unlabeled data to improve an object detector. However, thus far these methods have assumed that the unlabeled data does not contain out-of-distribution (OOD) classes, which is unrealistic with larger-scale unlabeled datasets. 
In this paper, we consider a more practical yet challenging problem, Open-Set Semi-Supervised Object Detection (OSSOD).
We first find the existing SSOD method obtains a lower performance gain in open-set conditions, and this is caused by the semantic expansion, where the distracting OOD objects are mispredicted as in-distribution pseudo-labels for the semi-supervised training.
To address this problem, we consider online and offline OOD detection modules, which are integrated with SSOD methods. 
With the extensive studies, we found that leveraging an offline OOD detector based on a self-supervised vision transformer performs favorably against online OOD detectors due to its robustness to the interference of pseudo-labeling.  
In the experiment, our proposed framework effectively addresses the semantic expansion issue and shows consistent improvements on many OSSOD benchmarks, including large-scale COCO-OpenImages.
We also verify the effectiveness of our framework under different OSSOD conditions, including varying numbers of in-distribution classes, different degrees of supervision, and different combinations of unlabeled sets.
\end{abstract}

\input{introduction}
\input{related_work}
\input{method}
\input{experiments}

\input{limitation}

\input{conclusion}


\noindent \textbf{Acknowledgments.}
We thank Ross Girshick for many helpful suggestions and valuable discussions.
Yen-Cheng Liu and Zsolt Kira were partly supported by DARPA’s Learning with Less Labels (LwLL) program under agreement HR0011-18-S-0044, as part of their affiliation with Georgia Tech.

\clearpage
%
%
\bibliographystyle{splncs04}
\bibliography{reference}

\clearpage
\input{appedix_section}
\end{document}

%% file: math_commands.tex
\usepackage{amsmath,amsfonts,bm}

\def\eqref#1{equation~\ref{#1}}

\def\1{\bm{1}}

\DeclareMathAlphabet{\mathsfit}{\encodingdefault}{\sfdefault}{m}{sl}
\SetMathAlphabet{\mathsfit}{bold}{\encodingdefault}{\sfdefault}{bx}{n}



%% file: introduction.tex
\section{Introduction}
\label{sec:intro}

The success of deep neural networks relies on large collections of labeled data, although creating such large-scale datasets is expensive and time-consuming. 
The recent development of successful Semi-Supervised Learning (SSL) methods alleviates this requirement by making better use of unlabeled data, narrowing the performance gap between the SSL models and the fully-supervised model.

Inspired by SSL methods for image classification~\cite{berthelot2019mixmatch,laine2016temporal,sajjadi2016regularization,tarvainen2017mean,zhang2018mixup,yun2019cutmix,guo2019mixup,hendrycks2020augmix,sohn2020fixmatch}, recent works on semi-supervised object detection (SSOD)~\cite{sohn2020simple,liu2021unbiased,zhou2021instant,tang2021humble} have applied the self-training method, which trains the model with the pseudo-labels of the unlabeled data. 
These works often consider a scenario where the labeled set is randomly sampled from a dataset (\textit{e.g.,} MS-COCO~\cite{lin2014microsoft}) and use the remaining images as the unlabeled set. This implicitly assumes the label spaces of labeled and unlabeled data are identical.
However, this closed-set assumption is unlikely to happen in real-world situations, where unlabeled images collected in the wild might contain out-of-distribution (OOD) objects, which are unseen, undefined, and unknown in the available labeled set.

\input{figure/teaser}

We are thus interested in a more practical yet challenging problem, Open-Set Semi-Supervised Object Detection (OSSOD), which aims to leverage the \textbf{unconstrained} unlabeled images (\textit{i.e.,} images containing unseen OOD objects), to improve an object detector trained with the available labeled data as shown in Figure~\ref{fig:teaser}\textcolor{red}{a}.
When adding unlabeled data containing open-set categories, we observe that the existing successful SSOD method leads to a lower performance gain or even \textit{degraded} results. 
This is different from the common belief that SSL methods can benefit from using more unlabeled data.
We attribute the above phenomena to the \opensetissue{} issue, where OOD objects are mispredicted as in-distribution objects with high confidence and misused as pseudo-labels with confidence thresholding (Figure~\ref{fig:teaser}\textcolor{red}{b}).

To eliminate the detrimental effect of OOD samples, we propose to add an additional OOD filtering process into the existing SSOD training pipeline.
More concretely, we first consider \textbf{online} OOD detectors to perform OOD filtering. 
An online OOD detector is a prediction head/branch we straightforwardly add on the object detector using existing OOD methods~\cite{hendrycks2019oe,lee2018simple,hsu2020generalized,tian2021geometric}.
However, we find that such methods cannot produce satisfactory results due to interference with other tasks, \textit{e.g.,} bounding box localization and box classification (See Section~\ref{sec:online_ood} for further discussion). 
In order to address this, we propose a simple but effective strategy that uses an \textbf{offline} OOD detection module, which is disentangled from the architecture of the object detector.
This OOD detector is based on a self-supervised DINO~\cite{caron2021emerging} model, and it provides several advantages.
First, the pre-training of DINO does not require label annotations, so it is suitable for the low-label setting and alleviates the concern of limited amounts of labels for OOD detection tasks.
Secondly, it is more effective in detecting OOD objects in the pseudo-labels compared with other (online) OOD methods as shown in Section~\ref{sec:OODDetection}, and this eliminates the detrimental effect of OOD samples for OSSOD tasks.
Lastly, since the architecture of OOD detector and object detector are independent, the training of the two models can be done separately and prevent interference as we observed in online OOD detectors.

\input{figure/goi_pred}

In our experiments, we first provide a systematic analysis between several OOD detection methods,
and our results suggest that offline methods show consistent improvements over the online detection methods. 
We also show that using the offline OOD detector can filter the OOD objects in pseudo-labels and consistently improve against the existing SSOD methods under different open-set scenarios, including under different combination of unlabeled sets, varying number of in-distribution (ID) classes, and different number of images.
We also find that using the pseudo-labels generated from our framework is even more effective than using the ground-truth labels provided in OpenImages (see Section~\ref{sec:ossod_expt}).

We highlight the contributions of this paper as follows:
\begin{itemize}
    \item To the best of our knowledge, we are the \textit{first} to address the open-set semi-supervised object detection tasks, and we analyze the limitation of existing SSOD methods on OSSOD tasks.
    \item We identify the challenges in designing OOD detection in OSSOD tasks, present online and offline OOD detectors, and provide a systematic comparison between these two modules.
    \item Through our extensive experiments, we demonstrate that an offline OOD detector can effectively remove OOD objects in pseudo-labels, and this leads to a significant improvements under different OSSOD scenarios, including a varying number of ID/OOD classes, different degrees of supervision, and different scales of datasets.
    \end{itemize}

%% file: figure/teaser.tex
\begin{figure}[t]
   \begin{picture}(0,105)
     \put(0,0){\includegraphics[width=0.5\linewidth]{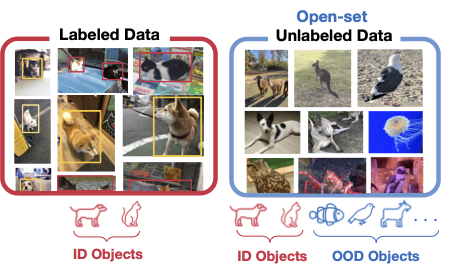}}
     \put(180,0){\includegraphics[width=0.48\linewidth]{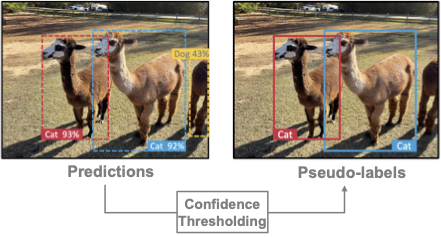}}
     \put(80,-8){(a)}
     \put(258,-8){(b)}
   \end{picture}
    \caption{(a) Open-Set Semi-Supervised Object Detection aims to learn an in-distribution object detector with a number of labeled images and another set of unconstrained/open-set unlabeled images. The objects appearing in the labeled data are defined as in-distribution (ID) objects, while some objects in the open-set unlabeled data are out-of-distribution (OOD) objects, which are unseen, unknown, and undefined in the labeled set. (b) While most of the self-training methods~\cite{sohn2020fixmatch,liu2021unbiased} rely on thresholding for filtering unreliable pseudo-labels, some OOD objects in unlabeled images are mispredicted as inlier objects with high confidence. These OOD objects are then mis-labeled, trained on, and lead to worse performance.
}
    \label{fig:teaser}
\end{figure}

%% file: figure/goi_pred.tex

%% file: related_work.tex
\section{Related Work}
\label{sec:related}

Open-set semi-supervised object detection focuses on improving an in-distribution object detector with in-the-wild unlabeled data and preventing the detrimental effects caused by distracting OOD objects in a more practical semi-supervised setup.
This task is different from the existing works on open-vocabulary/open-world object detection~\cite{zareian2021open,gu2021open,saito2021learning,zhou2022detecting,kim2022learning,joseph2021towards,huynh2021open}, where the goal is to improve the recall and accuracy of novel/OOD objects.

\noindent\textbf{Semi-Supervised Object Detection.}
Recent successful methods on semi-supervised learning for image classification have applied various data augmentation and consistency regularization on unlabeled data~\cite{berthelot2019mixmatch,laine2016temporal,sajjadi2016regularization,tarvainen2017mean,zhang2018mixup,yun2019cutmix,guo2019mixup,hendrycks2020augmix,sohn2020fixmatch}.
These techniques have also promoted the development of semi-supervised learning in object detection tasks. 
For example, several works~\cite{sohn2020simple,liu2021unbiased,zhou2021instant,tang2021humble} apply the self-training method, which is based on the Teacher-Student framework.
For example, STAC~\cite{sohn2020simple} uses the labeled data to train the Teacher, which generates pseudo-labels to supervise the Student model.
To refine the quality of pseudo-labels, existing works propose several techniques, including exponential moving average (EMA)~\cite{zhou2021instant,xu2021end,liu2021unbiased,yang2021interactive}, co-rectify mechanism~\cite{zhou2021instant}, replacing hard pseudo-labels with soft pseudo-labels~\cite{tang2021humble}, and addressing class imbalance in pseudo-labels~\cite{liu2021unbiased}.
While promising results have been made, existing SSOD works usually experiment on datasets where the labeled and unlabeled sets have the same object categories.
We are interested in a more challenging scenario, where some objects in the unlabeled set are novel and never appear in the labeled set. 

\noindent\textbf{Out-of-Distribution Detection.}
Out-of-distribution (OOD) classes are object categories that do not appear in the training label space and are not known \textit{a-priori}, and OOD detection, which plays a vital role in OSSOD tasks, is a binary classification problem that decides whether a sample is an ID or OOD.
Early works use different scoring functions to estimate the likelihood of a sample being OOD, including maximum softmax probability (MSP)~\cite{hendrycks2017baseline,liang2018odin} and Mahalanobis distance~\cite{lee2018simple}.
Another line of works~\cite{hendrycks2019oe,liu2021energy,mohseni2020self} exploit large sets of OOD samples during training and they generally achieve better performance on detecting the outlier objects.
There are also several generative-based methods~\cite{zong2018deep,pidhorskyi2018generative,sabokrou2018adversarially,nalisnick2018deep} for OOD detection tasks, though it is difficult and prohibitively challenging to apply them on large-scale semi-supervised object detection tasks.  
Among the existing OOD works, Fort \textit{et al.}~\cite{fort2021exploring} shows state-of-the-art performance on OOD detection by simply fine-tuning the pretrained vision transformer with the available inlier data.
While promising results have been made on OOD detection tasks, most of the existing works only experiment on small-scale image classification tasks, and their extension to the large-scale object detection tasks is not verified.
The only work on OOD detection for object detection is VOS~\cite{du2022vos}, but the interference with pseudo-labeling in SSOD tasks and computation cost of virtual negative sampling make it incompatible with the existing SSOD methods. 

\noindent\textbf{Open-Set Semi-Supervised Learning.}
Existing works on open-set semi-supervised learning focus on image classification tasks~\cite{yu2020multi,saito2021openmatch,huang2021trash,luo2021consistency} or image generation~\cite{girish2021towards}. 
For example, MTC~\cite{yu2020multi} estimates the OOD score for each unlabeled image, and both the network parameters and OOD scores are updated alternately. 
Based on the estimated OOD scores, MTC eliminates the OOD samples with low OOD scores and improves the image classifier in a semi-supervised scenario.
OpenMatch~\cite{saito2021openmatch} applies a consistency regularization on a one-vs-all classifier, which is used as an OOD detector to filter the OOD samples during semi-supervised learning. 
Despite the promising results, no prior work has addressed open-set semi-supervised learning for object detection tasks, which have more challenges than image classification tasks. 
For instance, in image classification, each image only contains one single object/class, whereas, in object detection, each image may contain an arbitrary number of ID/OOD objects, making pseudo-labeling integrated with OOD detection much more challenging. 
Additionally, we also observed that balancing object detection losses against OOD detector is difficult as we showed in Section~\ref{sec:online_ood}.

%% file: method.tex
\section{Revisiting Semi-Supervised Object Detection}
Semi-supervised object detection (SSOD) aims to learn an object detector by using a set of labeled images $\bm{D}_s =\{\bm{x}^s_i, \bm{y}^s_i\}^{N_s}_{i=1}$ and  unlabeled images $\bm{D}_u=\{\bm{x}^u_i\}^{N_u}_{i=1}$, where $\bm{N}_s<<\bm{N}_u$.
Existing SSOD works~\cite{sohn2020simple,liu2021unbiased,zhou2021instant,xu2021end} apply the self-training method and have shown significant improvements in this semi-supervised scenario.
The typical strategy of such methods is to generate pseudo-labels (\textit{i.e.,} pseudo-boxes and the corresponding class labels) of unlabeled images and then train the object detector using both labeled data and unlabeled data with pseudo-labels.

These works adopt the Teacher-Student framework, where the \textbf{Teacher} model generates pseudo-labels to train the \textbf{Student} model.
Specifically, the \textbf{Teacher} model takes the weakly-augmented unlabeled data as input and generates the box predictions, and the bounding boxes with confidence higher than a pre-defined threshold $\tau$ are selected as the pseudo-boxes. 
The \textbf{Student} then takes as input the same images with stronger augmentation, and we enforce the consistency loss $\mathcal{L}_{unsup}$ (\textit{i.e.,} unsupervised loss) between its predictions and the generated pseudo-labels.  
To train the \textbf{Student}, we combine both the supervised loss $\mathcal{L}_{sup}$ and the unsupervised loss $\mathcal{L}_{unsup}$.
\begin{equation}
\begin{split}
\mathcal{L}_{ssod} &= \mathcal{L}_{sup}+\lambda\mathcal{L}_{unsup} =  \sum_i     \mathcal{L}(\bm{x}^s_i, \bm{y}^s_i) + \lambda \sum_i     \mathcal{L}(\bm{x}^u_i, \bm{\hat{y}}^u_i),  \\
\end{split}
\end{equation}\label{eq:loss_ssod}
where $\lambda$ is the unsupervised loss weight and $\bm{\hat{y}}^u_i=\delta( \bm{\tilde{y}}^u_i; \tau)$ represents the pseudo-labels, which are derived from the bounding box prediction $\bm{\tilde{y}}^u_i$ after the confidence thresholding function $\delta(\cdot)$ with the pre-defined threshold $\tau$.

To further refine the quality of pseudo-labels, some existing works~\cite{tang2021humble,liu2021unbiased,xu2021end} update the \textbf{Teacher} model ($\theta_t$) by exploiting the model weights of \textbf{Student} ($\theta_s$) via the exponential moving average (\textit{i.e.,} $\theta_{t} \leftarrow \alpha \theta_{t} + (1-\alpha) \theta_{s}$).

To verify the effectiveness of the models, existing works often experiment on a setting where the labeled and unlabeled sets are randomly sampled from the same dataset.
This implicitly assumes the set of object categories are the same across the labeled and unlabeled sets, and we formalize this setup as the \textbf{closed-set} SSOD.

\section{Open-Set Semi-Supervised Object Detection}
While the above self-training methods show promising results in closed-set SSOD, it restricts the set of object categories in the unlabeled set to be the same as the labeled set.
This setting, however, is less practical since unlabeled datasets collected in the wild might contain images with novel objects beyond what were presented in the labeled dataset.
We are thus interested in OSSOD, a more practical yet challenging problem setup, where novel and undefined objects appear in the unconstrained unlabeled set and are not available in the labeled set with a limited amount of data. 

\subsection{Semantic Expansion in Open-Set Scenarios}\label{sec:semanitc_exp}

\input{figure/openset_issue}

\input{table/compare-ssod}

We first analyze the existing successful SSOD work\footnote{ \scriptsize As all SoTA SSOD methods~\cite{liu2021unbiased,tang2021humble,zhou2021instant,xu2021end} use the Teacher-student mechanism and pseudo-labeling method, we choose UT~\cite{liu2021unbiased} as an example while our framework is not restricted to it.} on OSSOD tasks. Interestingly, we found it shows limited performance gain in the open-set scenario compared with closed-set scenario as shown in Figure~\ref{fig:compare_ssod}.
We attribute this lower performance gain to \textbf{\opensetissue}\, as presented in Figure~\ref{fig:sem_exp}\textcolor{red}{a}.
To be more specific, when OOD/novel objects exist in the unlabeled set, the closed-set classifier (\textit{e.g.,} ROIhead classifier in Faster-RCNN~\cite{ren2015faster}) might mispredict these OOD instances as ID objects with high confidence (a similar observation is also made recently in prior works~\cite{dhamija2020overlooked,miller2021uncertainty}).
These over-confident and incorrect OOD instances are prone to be selected as the pseudo-labels of ID objects even after confidence thresholding, and using these incorrect OOD pseudo-labels in the training makes the EMA-updated {Teacher} mispredict more OOD objects as ID objects.
After the iterative EMA updates between the Teacher and Student, the semantic discrimination of ID objects is \textit{enlarged} and incorrectly covers more OOD objects. 
This causes the false-positive rate of the OOD pseudo-labels to increase as the model is trained longer, as presented in Figure~\ref{fig:sem_exp}\textcolor{red}{b}.

To address the above issue and improve the performance on OSSOD tasks, we propose an integrated framework to detect the OOD instances in the pseudo-boxes and eliminate their detrimental effect during semi-supervised learning.
With the goal of deciding whether a sample is an ID or OOD data (\textit{i.e.,} binary classification), prior works have proposed several OOD detection methods~\cite{hendrycks2017baseline,lee2018simple,liu2021energy,hendrycks2019oe,saito2021openmatch}, although most of these works only verify the effectiveness of the methods on the image-level vision tasks. 
To further advance to OOD detection for instance-level tasks (\textit{e.g.,} object detection), we consider \textbf{online} and \textbf{offline} OOD detectors in the following sections.

\subsection{Online OOD detection}\label{sec:online_ood}
\input{figure/framework}
\input{table/dynamic_pseudolabel}

\input{table/with_without_OE}

As illustrated in Figure~\ref{fig:ood_filter}, online OOD detection expands the architecture of the existing object detector so that it can perform OOD detection in addition to bounding box classification and localization.
One simple way to detect the OOD objects is to use the maximum softmax probability (MSP)~\cite{hendrycks2017baseline} of ROIhead classifier as an indicator for identifying OOD samples.
In a similar spirit, another way is to apply a range of existing OOD detection methods~\cite{hendrycks2017baseline,lee2018simple,liu2021energy,hendrycks2019oe,saito2021openmatch,tian2021geometric} and add an additional OOD branch on the ROIhead.
We combine the semi-supervised object detection loss $\mathcal{L}_{ssod}$ as mentioned in Eq.~\ref{eq:loss_ssod} and OOD detection loss $\mathcal{L}_{ood}$ defined in the original works, and we train the model in an end-to-end manner.

However, these online OOD detectors have several limitations in OSSOD tasks. 
On the one hand, as presented in Figure~\ref{fig:dynamic_pseudo}\textcolor{red}{a}, pseudo-labeling used for SSOD significantly degrades the performance of online OOD detection (we use MSP~\cite{hendrycks2017baseline} as an example), and such a trend is hypothetically caused by the instability of pseudo-labels (\textit{i.e.,} classification noise in pseudo-labels).
On the other hand, bundling two different tasks (\textit{e.g.,} OOD detection and object detection) in a shared architecture leads to sub-optimal performance of object detection task as shown in Figure~\ref{fig:dynamic_pseudo}\textcolor{red}{b}, making the online OOD detection less favorable to be applied for OSSOD tasks.

\subsection{Offline OOD detection.}\label{sec:offline_ood}

To ameliorate the above issues, as shown in Figure~\ref{fig:ood_filter}, we alternatively propose to use \textbf{offline} OOD detection, where we construct an OOD detector that is disentangled from the architecture of the object detector and \textit{not} jointly trained with the object detector.
This framework design provides two advantages:
1) the offline OOD detector is compatible with \textit{any} existing SSOD methods~\cite{sohn2020simple,zhou2021instant,liu2021unbiased,zhu2019soft}, since the offline OOD detector is modularized and independent from the object detector. 
2) Such a framework design alleviates the concern about competing task objectives between OOD detection and object detection.

To design the offline OOD detector, we exploit the current state-of-the-art OOD method, which is simple yet effective on image-level OOD tasks.
Specifically, as pointed out by Fort \textit{et al.}~\cite{fort2021exploring}, one highly successful method for OOD detection is to simply fine-tune a pretrained ViT-B~\cite{dosovitskiy2020image} with available ID data.
Despite its promising results, pre-training a ViT-B requires ImageNet-21k images and the corresponding large amount of annotated labels, which is not suitable for our semi-supervised setting and would result in an unfair comparison to prior works.

\noindent\textbf{DINO as an offline OOD detector.}
We therefore utilize the self-supervised DINO~\cite{caron2021emerging}, which is only trained with ImageNet-1k images without using any label annotations\footnote{ \scriptsize The backbone of object detectors is usually pretrained on the ImageNet1k classification~\cite{wu2019detectron2,mmdetection}, and both images and ground-truth labels are required for pretraining of the model weights. A systematic comparison between using DINO and ViT will be discussed in supplementary material.}.
To perform the OOD detection, we fine-tune DINO with multi-class training and then compute different OOD scores (described below) to decide whether an object is an ID or an OOD sample.
To be more concrete, the OOD detector is trained as a simple classifier with $K+1$ output classes ($K$ is the number of ID object classes), where the additional node is for the OOD and novel samples.
Unlike prior works~\cite{mohseni2020self,hendrycks2019oe,liu2021energy} that used a large amount of additional OOD data during training, we do not assume the availability of such data. We therefore only use the available ID labeled data and propose to regard the background proposal patches as the OOD samples.
Specifically, we crop the image patches according to the ground-truth boxes, and the background patches are randomly sampled from the proposal boxes labeled with the low IoU to the ground-truth boxes. 
We compute the OOD detection loss $\mathcal{L}_{ood}$ with the cross-entropy which enforces the ID patches as the corresponding ID foreground labels and enforces the background patches as the OOD class. 
While it is sub-optimal to use background instances as OOD class, they still provide sufficient negative gradients for distinguishing the ID objects when the novel OOD objects are not available in the labeled set.

After the fine-tuning of the DINO model, we freeze the DINO model and experiment
the following common scores for OOD detection:

\noindent 1. \textbf{Mahalanobis Distance~\cite{lee2018simple}}:
\begin{equation}\label{eq:mahalanobis_distance}
\gamma_{ood} = \max_k- (f(x) - \tilde{\mu}_k)^{\intercal} \tilde{\Sigma}^{-1} (f(x) - \tilde{\mu}_k).
\end{equation}
where $f(x)$ represents the DINO intermediate feature vector of patch $x$, $\tilde{\mu}_k$ is the estimated mean vector of class $k$, and $\tilde{\Sigma}$ is the estimated covariance matrix. Both $\tilde{\mu}_k$ and $\tilde{\Sigma}$ are estimated by using the available labeled data.

\noindent 2. \textbf{Inverse Abstaining Confidence}~\cite{mohseni2020self,thulasidasan2021effective}:
\begin{equation}\label{eq:inverse_abstaining_confidence}
\gamma_{ood} = 1 - \tilde{p}_{K+1},
\end{equation}
where $\tilde{p}_{K+1}$ is the prediction confidence of $K+1$-th class (\textit{i.e.,} abstention class) from the DINO classifier. 

\noindent 3. \textbf{Energy Score}~\cite{liu2021energy}:
\begin{equation}\label{eq:energy_score}
\gamma_{ood} = - T \log \sum^K_i e^{f_i(x)/T},
\end{equation}
where $T$ is the temperature value, $K$ is the number of classes, $f_i(x)$ represents the $i$-th index of the logit corresponding to the class $i$.
We also consider \textbf{Shannon Entropy} and \textbf{Euclidean Distance} for the OOD scores.
To decide whether an object is an ID or OOD class, we set a threshold $\delta_{ood}$ on the OOD score, and the objects with OOD scores lower than the threshold are regarded as OOD samples.

\noindent\textbf{Integration with Semi-Supervised Object Detection.}
With the goal of filtering the OOD objects in pseudo-labels, we integrate the trained OOD detector with the SSOD method by inserting the OOD filtering after the confidence thresholding as shown in Figure~\ref{fig:ood_filter}. 
In other words, the pseudo-boxes are derived by sequentially applying both confidence thresholding and OOD filtering on predicted boxes from Teacher model, and they are then used to compute the unsupervised loss defined in Eq.~\ref{eq:loss_ssod} to train the Student.
Note that our OOD detector is complementary to the existing semi-supervised object detection works~\cite{sohn2020simple,zhou2021instant,liu2021unbiased,zhu2019soft} and can also be combined with these SSOD methods to address the open-set issue.

%% file: figure/openset_issue.tex
\begin{figure*}[t]
   \begin{picture}(0,95)
     \put(0,0){\includegraphics[width=0.8\linewidth]{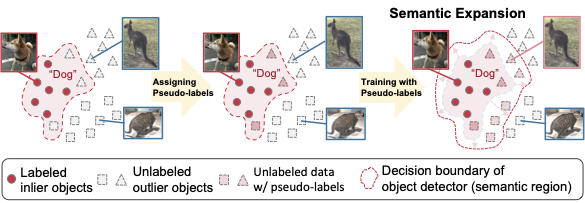}}
     \put(280,0){\includegraphics[width=0.2\linewidth]{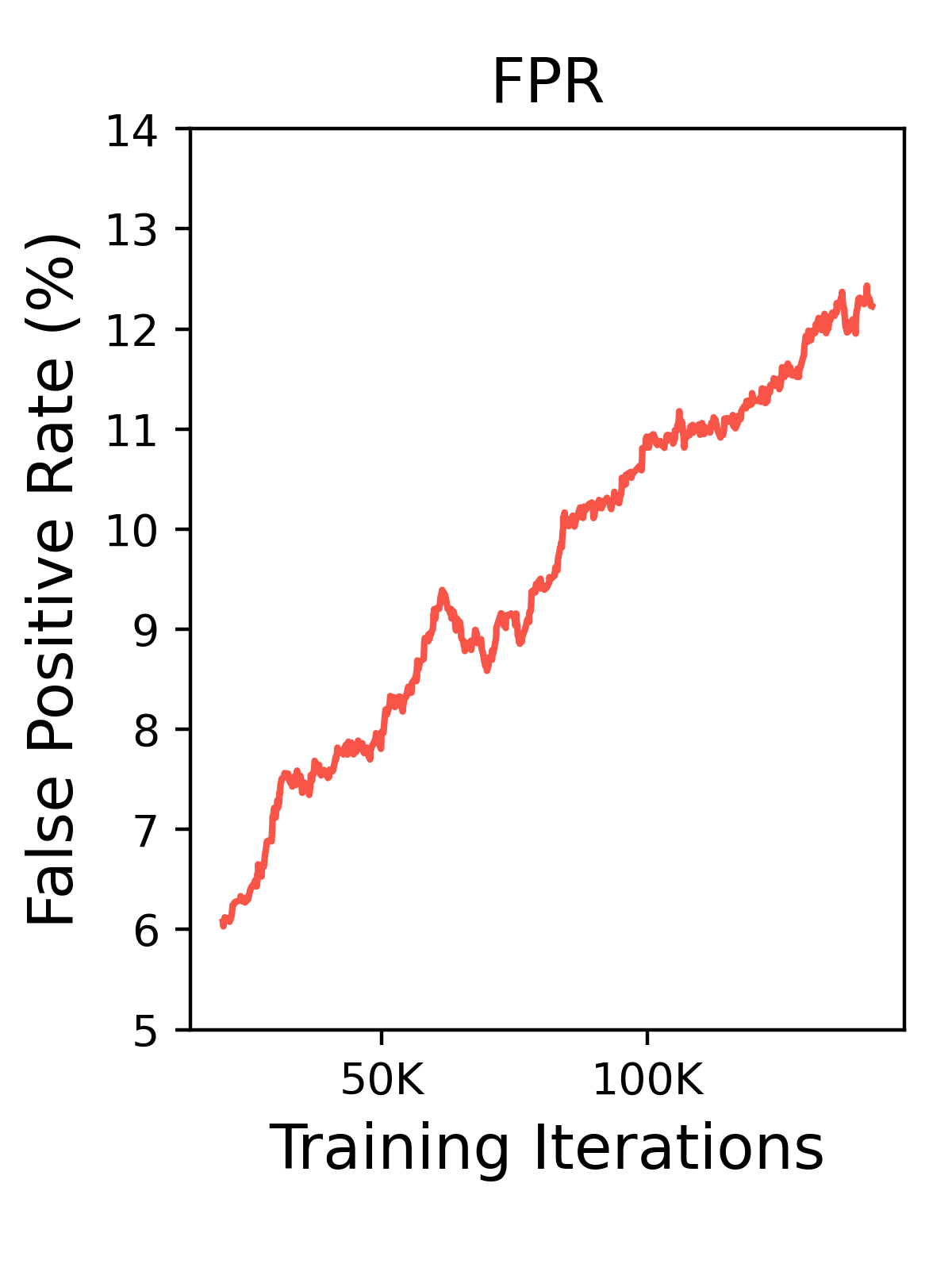}}
     \put(138,-10){(a)}
     \put(313,-10){(b)}
   \end{picture}
    \caption{Illustration of \textbf{semantic expansion}. (a) With the object detector trained with the labeled data, some OOD objects in unlabeled data are predicted as ID objects, and solely using the confidence thresholding based on the box score cannot effectively suppress these OOD objects in pseudo-labels. Using the noisy pseudo-labels in self-training methods makes the open-set issue become more severe after several training iterations, and (b) thus the false-positive rate (\textit{i.e.,} percentage of OOD objects predicted as ID) increases over time. }
    \label{fig:sem_exp}
\end{figure*}

%% file: table/compare-ssod.tex
\begin{wrapfigure}{r}{0.45\textwidth}
\begin{center}
\includegraphics[width=\linewidth]{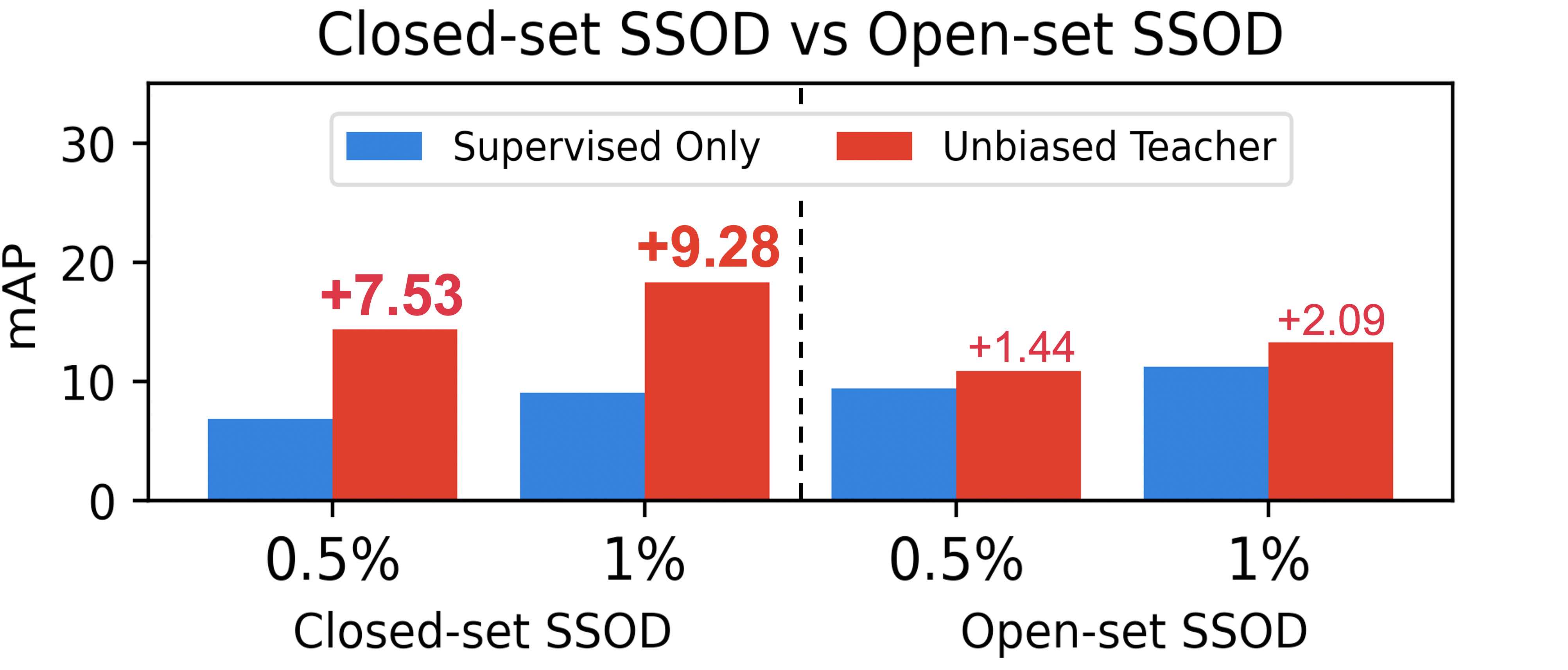}
\end{center}
\caption{Performance comparison between closed-set and open-set SSOD. The performance gain of the existing SSOD method~\cite{liu2021unbiased} is much smaller in the open-set conditions. Note that we randomly select $0.5\%/1\%$ data as the labeled set for both conditions. 
}
 \label{fig:compare_ssod}
\end{wrapfigure}

%% file: figure/framework.tex
\begin{figure*}[t]
   \begin{picture}(0,140)
   \centering
     \put(0,-10){\includegraphics[width=1.0\linewidth]{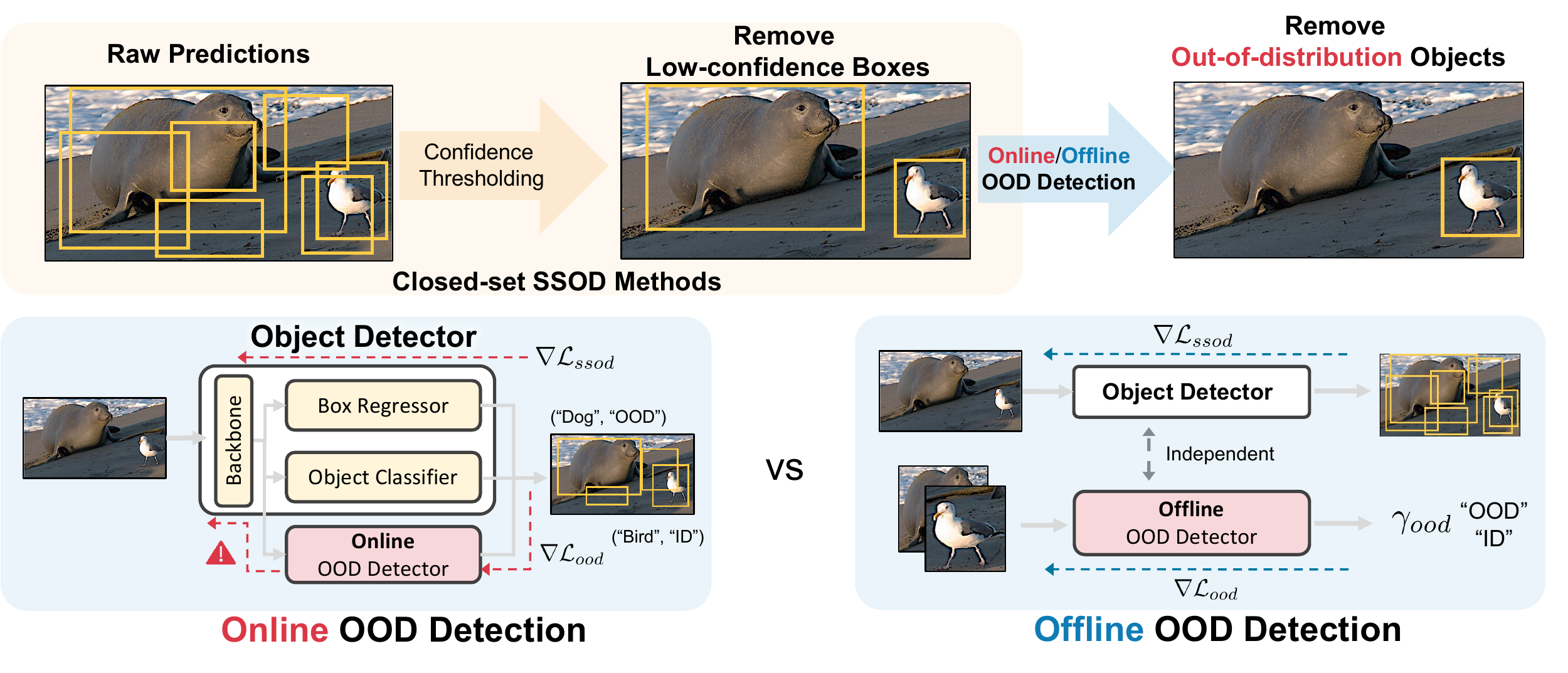}}
   \end{picture}
    \caption{
    Illustration of online and offline OOD detection frameworks for OSSOD tasks. Combined with the existing closed-set SSOD methods, OOD detection aims to remove the OOD objects in pseudo-labels and prevent \opensetissue{} in OSSOD.
    For the online OOD detector, we add an additional branch on the object detector, but it suffers from the instability of pseudo-labels used for SSOD and leads to degraded performance of the object detector. In constrast, the offline OOD detector does not have these limitations and produce better results on OOD detection due to its nature of independent model architecture.
    }
    \label{fig:ood_filter}
\end{figure*}

%% file: table/dynamic_pseudolabel.tex
\begin{figure*}[t]
   \begin{picture}(0,80)
   \centering
     \put(20,5){\includegraphics[width=0.5\linewidth]{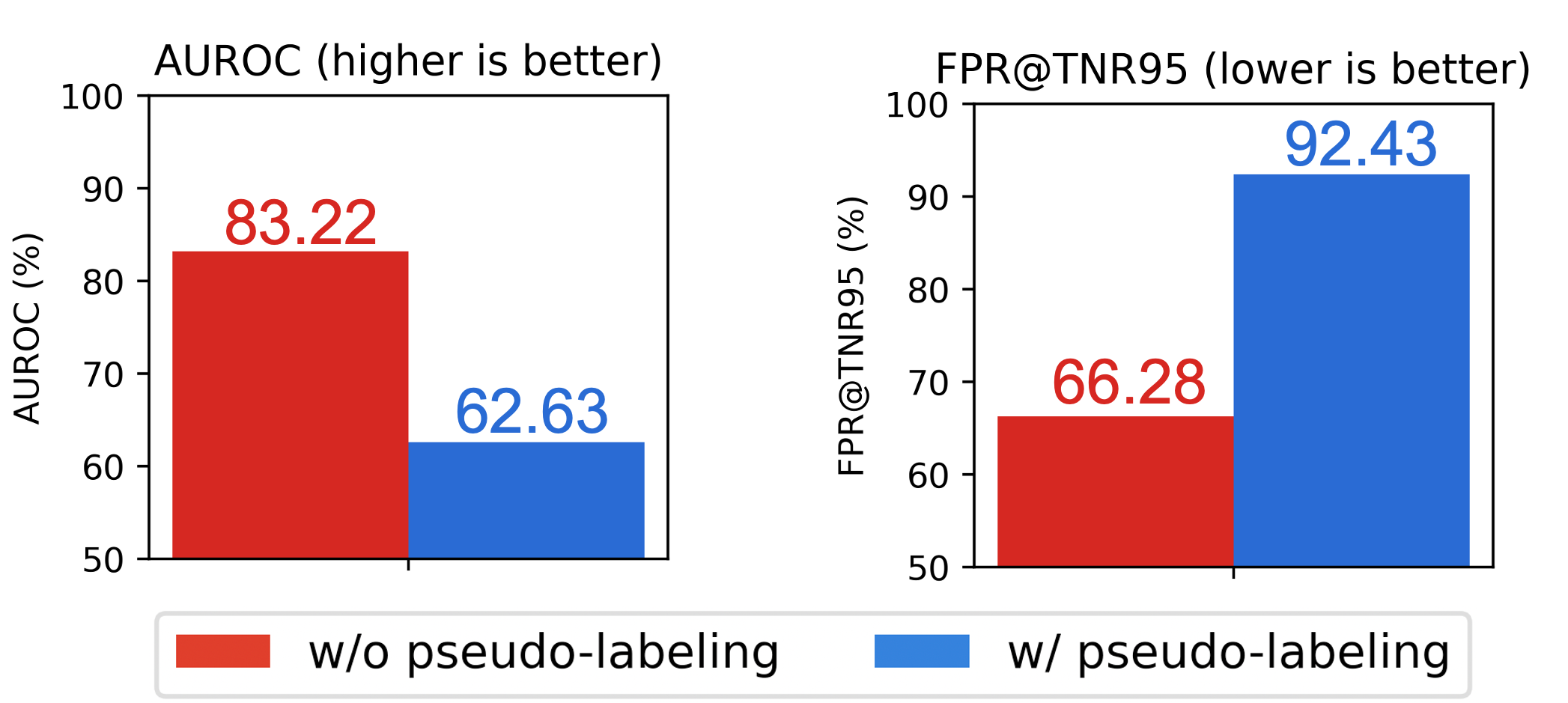}}
     \put(195,4){\includegraphics[width=0.39\linewidth]{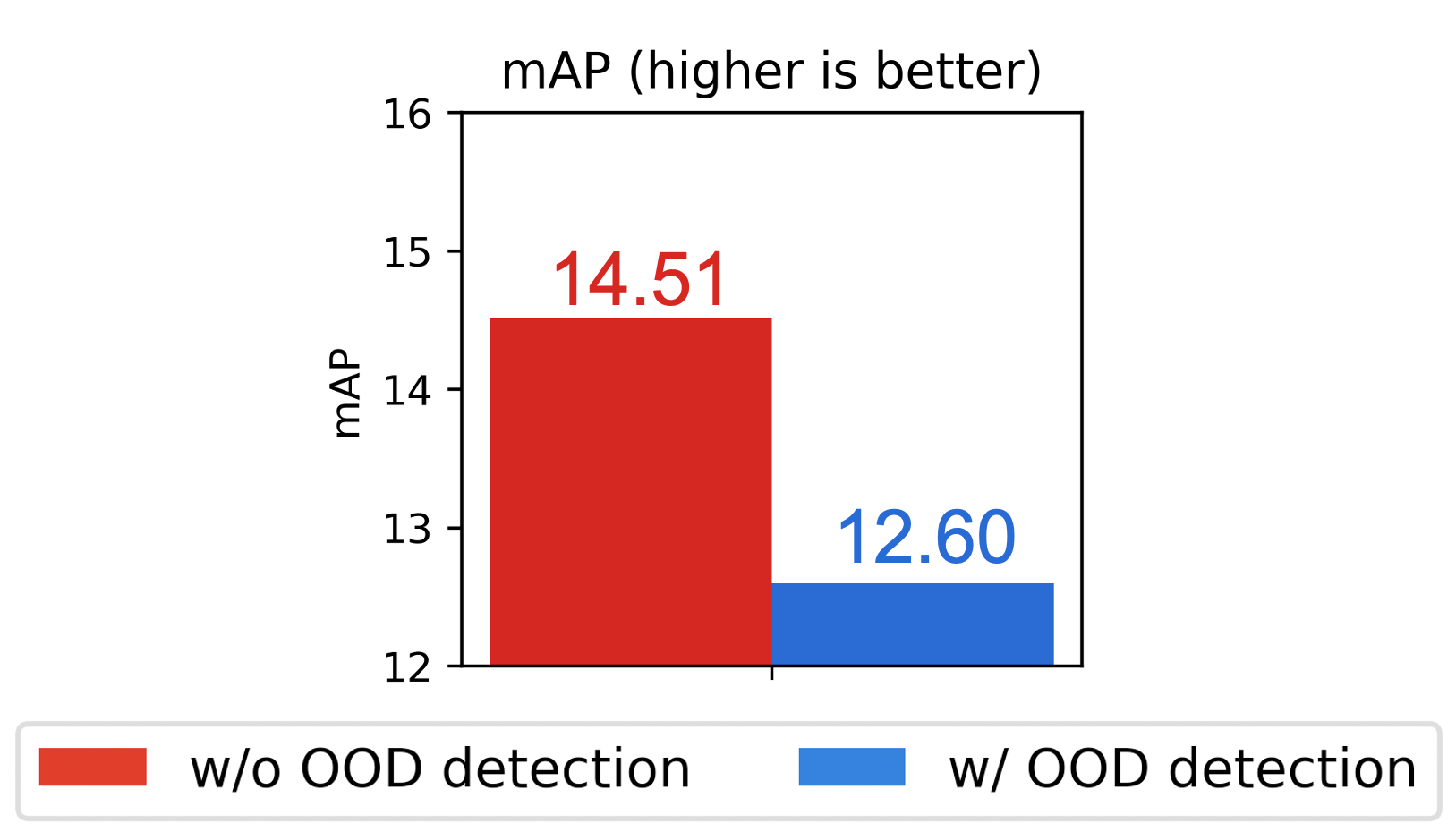}}

     \put(105,-5){(a)}
     \put(262,-5){(b)}
   \end{picture}
\caption{
{(a) Pseudo-labeling used in SSOD degrades the performance of online OOD detection.} 
{(b) Accuracy of an object detector is degraded by adding online OOD detection}, where we create another branch on ROIhead and exploit outlier exposure~\cite{hendrycks2019oe} to detect OOD samples. 
Both AUROC and FPR@TNR95 are standard evaluation metrics for OOD detection (defined in Section~\ref{sec:compare_onoff}).}
    \label{fig:dynamic_pseudo}
\end{figure*}

%% file: table/with_without_OE.tex

%% file: experiments.tex
\section{Experiments}\label{sec:expt}

\subsection{Experimental Setting and Datasets}

\noindent\textbf{COCO-Open.}
\texttt{COCO2017-train} contains $117$k images with box-level labels from $80$ object categories.
We randomly sample $20$/$40$/$60$ classes as the ID classes and the remaining classes as the OOD classes.
Specifically, since each image contains multiple objects, \texttt{COCO2017-train} is divided into the pure-ID, mixed, and pure-OOD image sets. 
Each image in the pure-ID set has at least one ID object and no OOD object, and each image in the pure-OOD set has at least one OOD object and no ID object.
As for the mixed set, each image has at least one ID object and at least one OOD object.
To make sure that OOD objects are not in the labeled set, we randomly sample the pure-ID images as a labeled set and the remaining data as the unlabeled set. Thus, the unlabeled set contains images from the pure-ID, pure-OOD, and mixed sets.


\noindent\textbf{COCO-OpenImages.}
To examine all methods in a large-scale OSSOD experiment, we use \texttt{COCO2017-train} as the labeled set and \texttt{OpenImagesv5}~\cite{OpenImages2} as the unlableed set.
\texttt{OpenImagesv5} contains 1.7M images with 601 object categories, and most object categories do not exist in COCO (and are hence OOD). 

We also include more experiments in the supplementary material.

 \subsection{Comparison between online and offline OOD detection}\label{sec:OODDetection}\label{sec:compare_onoff}

We first present a comparison between online and offline detectors on \textbf{OOD detection} tasks (object detection results shown in Section~\ref{sec:ossod_expt}), and we evaluate them with the standard OOD detection metrics, including area under the ROC curve (AUROC) and false-positive rate under the true-false rate with $x\%$ (FPR@TNR$x$). 
We consider several OOD detection methods~\cite{hendrycks2017baseline,lee2018simple,liu2021energy,hendrycks2019oe,saito2021openmatch,saito2021openmatch,hsu2020generalized,tian2021geometric}\footnote{\scriptsize As the problem setting of GODIN~\cite{hsu2020generalized} and GSD~\cite{tian2021geometric} is OOD detection without OOD samples, we adapt them to our setup and enhance the performance by using an linear classifier to predict background samples.}.
For a fair and thorough comparison, we adapt these methods from image classification to object detection tasks and implement them in a unified framework with the same implementation details such as batch size and optimization. 
To better understand the efficacy of each OOD method and prevent the interference of pseudo-labeling from affecting the performance, we do not use the pseudo-labels of unlabeled data and only use the available labeled data to train all OOD detectors (adding pseudo-labels worsen online OOD detection methods as we discussed in Section~\ref{sec:online_ood}).

 \input{table/coco-open-ood}

As presented in Table~\ref{table:ood_detection}, we list our observations as follows, 

1) \textit{offline OOD detectors are significantly better than online OOD detectors on all evaluation metrics}, and this trend also supports that the disentanglement between OOD detector and object detector can alleviate the interference between OOD detection and object detection. 
It is worth repeating that the feature backbone of Faster-RCNN is pretrained on ImageNet1k with label annotations, so the superiority of the DINO on OOD detection does not come from using extra images and labels. 
In addition, as mentioned in Section~\ref{sec:offline_ood}, another merit of the offline OOD detector is its robustness to the instability to pseudo-labeling used in SSOD due to the isolated training of OOD detectors and object detectors.  

2) Among the OOD scores used in offline detectors, simply applying the Euclidean distance performs the bests on all metrics. 
The popular Mahalanobis distance cannot lead to satisfactory results due to the inaccurate covariance matrix estimated limited amounts of labeled data.

Additionally, in the supplementary material, we also list other in-depth analyses and discussions on OOD detection, including using supervised ViT-B as an offline OOD detector, and other offline OOD detection methods with different ID/OOD class splits and different degrees of supervision.
Owing to the superior performance on OOD detection, we apply the offline OOD detectors in the following OSSOD experiments.

\subsection{Experiments on Open-set Semi-Supervised Object Detection}\label{sec:ossod_expt}

In addition to the OOD detection performance shown in the previous section, we further advance to OSSOD tasks and assess the object detection performance under different OSSOD conditions, including different combinations of unlabeled sets,  varying numbers of ID classes, different degrees of supervision, and larger-scale OSSOD tasks (COCO-additional and COCO-GOI). 

\input{figure/coco-open-datasource}

\noindent\textbf{Effect of different combinations of unlabeled sets.}
We experiment with the existing SSOD method (UT~\cite{liu2021unbiased}) and our proposed OOD detectors by using different combinations of unlabeled data, including the pure-ID, the mixed, and the pure-OOD sets.
The purpose of this experiment is to verify the existence of open-set issue/semantic expansion and investigate whether the OOD samples affect the SSOD methods. 

As shown in Figure~\ref{fig:datasource}, UT achieves $22.10$ mAP when the pure-ID images are used in the unlabeled set, while the performance drops when the mixed and pure-OOD sets, containing OOD objects, are added. 
This shows that OOD objects in the unlabeled set do affect the effectiveness of the existing SS-OD method, and the trend is also contrary to the common belief that semi-supervised methods can derive more performance gains by using more unlabeled data. 
Therefore, with our proposed OOD filtering (OF-DINO), the performance of the object detector does not degrade but instead improves by using the mixed set, and this reflects the importance of eliminating the detrimental effect of OOD objects. 

\noindent\textbf{Varying numbers of ID classes.}
We also experiment by varying the number of ID classes in COCO-Open, and we first find that UT shows less performance gain against label-only baseline when there are fewer ID classes (see Table~\ref{table:coco-open-cls}).
This is because the semantic expansion/open-set issue becomes more severe when more OOD objects appear in the unlabeled set, and such an issue can be alleviated by the OOD filtering.
With the OOD filtering, we can obtain a larger improvement gain against UT, and it improves $6.54$ mAP against the labeled-only baseline and $5.06$ mAP against UT in the case of $20$ ID classes.
This verifies the efficacy of the OOD filtering using the different number of ID/OOD classes.

\input{table/coco-open}
\input{table/coco-open-num-img}

\noindent\textbf{Different degrees of supervision.}
In addition, we consider varying the number of labeled images in COCO-Open as shown in Table~\ref{table:coco-open-numimg}, and we find that using the OOD detector can also consistently improve against the label-only baseline and UT under different numbers of labeled images. 
When fewer labeled images are used, we could obtain more improvement gain by filtering OOD samples.

\noindent\textbf{Analysis on pseudo-labels.}
To better understand how our OOD detector helps the learning of the SSOD method in the open-set condition, we use the ground-truth labels to measure the number of ID/OOD boxes in pseudo-labels (note that we only use ground-truth labels for the analysis and do not use them in the training).
As illustrated in Figure~\ref{fig:analysis_box}, our OOD detector can effectively suppress OOD objects in pseudo-labels without sacrificing the ID objects, and thus this alleviates the semantic expansion issue mentioned in Section~\ref{sec:semanitc_exp} and leads to improvement for the OSSOD task.

\input{figure/analysis_fpr}

\noindent\textbf{COCO-GOI.}
We also consider a more challenging experiment, where we use unlabeled \texttt{OpenImagesv5} to improve the object detector trained on \texttt{COCO2017-train}.
Although \texttt{OpenImagesv5} is substantially larger than COCO datasets, the object categories in \texttt{OpenImagesv5} (601 classes) are more diverse than the object categories in MSCOCO (80 classes).
This implies there are more distracting OOD objects that can affect the performance of existing SSOD methods. 
In addition to UT, we also consider a fully-supervised baseline, which uses ground-truth labels from \textit{both} \texttt{COCO2017-train} and \texttt{OpenImagesv5} to train an object detector. 
To be more specific, we first manually label the correspondence between 601 classes in \texttt{OpenImagesv5} and 80 classes in \texttt{COCO2017-train}, and we only use labels of 80 COCO-classes in \texttt{OpenImagesv5}. 
We present the details of the correspondence in our supplementary material. 

As presented in Table~\ref{tab:coco_goi}, with our OOD filtering mechanism (DINO) we can improve UT from $41.81$ mAP to $43.14$ mAP.
Another interesting result is that our framework can perform \textit{even better than the model with the ground-truth labels from OpenImages}, and this indicates that using pseudo-labels generated from our framework might be more effective than the ground-truth labels provided in \texttt{OpenImages}. This is potentially caused by noisy labels from human annotations.
In addition, we also find that some COCO object categories are rare in \texttt{OpenImages}, and this potentially limits further improvements by exploiting \texttt{OpenImages} as an unlabeled set (more discussion in the supplementary material).

\input{table/coco-goi}

%% file: table/coco-open-ood.tex

\begin{table}[t]
\centering
\renewcommand{\arraystretch}{1.3}
\caption{Evaluation of OOD detection for object detection. We sample $20$ classes from COCO as ID objects, and  $4k$ pure-ID images are selected as labeled images. 
}
\label{table:ood_detection}
\resizebox{0.6\linewidth}{!}{%
\begin{tabular}{cccccc}
\toprule
OOD Models                        & Methods                    & OoD Scores $\gamma_{ood}$    & AUROC$\uparrow$ & FPR75$\downarrow$ & FPR95$\downarrow$ \\

\midrule
\multirow{9}{*}{\begin{tabular}[c]{@{}c@{}}Online \\ (ROIhead)\end{tabular}} & \multirow{5}{*}{Vanilla}               & MSP~\cite{hendrycks2017baseline}   & 67.0               & 58.4               & 92.3                \\
                                                  &                                        & Energy~\cite{liu2021energy}      & 75.5                     & 36.8                & 83.6            \\
                                                  &                                        & Entropy     & 75.9               & 38.5               & 83.1            \\
                                                  &                                        & Mahalanobis~\cite{lee2018simple} & 50.2                     & 83.0               & 98.1             \\
                                                  &                                        & Euclidean   & 56.3                  & 74.3            & 96.1              \\
                                                  & OE~\cite{hendrycks2019oe}& MSP         & 67.0            & 55.0               & 89.1                \\
                                                  & OVA~\cite{saito2021openmatch}   & MSP         & 73.0                     & 45.7               & 90.0              \\
                                                  & GODIN~\cite{hsu2020generalized}        & Cosine $h(x)$  & 77.8                      & 33.8              & 77.4                \\
                                                  & GSD~\cite{tian2021geometric}           & Feat. angle & 78.7                      & 32.1            & 73.9          \\ \midrule
\multirow{5}{*}{\begin{tabular}[c]{@{}c@{}}Offline \\ (DINO)\end{tabular}}                             & \multirow{5}{*}{ Ours }  & IAC~\cite{thulasidasan2021effective}     & 83.6                  & 22.4             & 61.7               \\
                                                  &                                        & Energy      & 89.6                     & 12.2             & 47.5              \\
                                                  &                                        & Entropy     & 88.9                  & 12.6               & 51.1               \\
                                                  &                                        & Mahalanobis~\cite{lee2018simple} & 81.8                   & 25.6              & 57.6               \\
                                                  &                                        & Euclidean   & \textbf{90.8}                 & \textbf{10.7}              & \textbf{38.6}            \\ 
              \bottomrule
\end{tabular}
}
\end{table}

%% file: figure/coco-open-datasource.tex
\begin{figure}[!!t]
\begin{center}
\begin{minipage}[t]{0.45\linewidth}
\raisebox{-3.2cm}
{\includegraphics[width=0.98\linewidth]{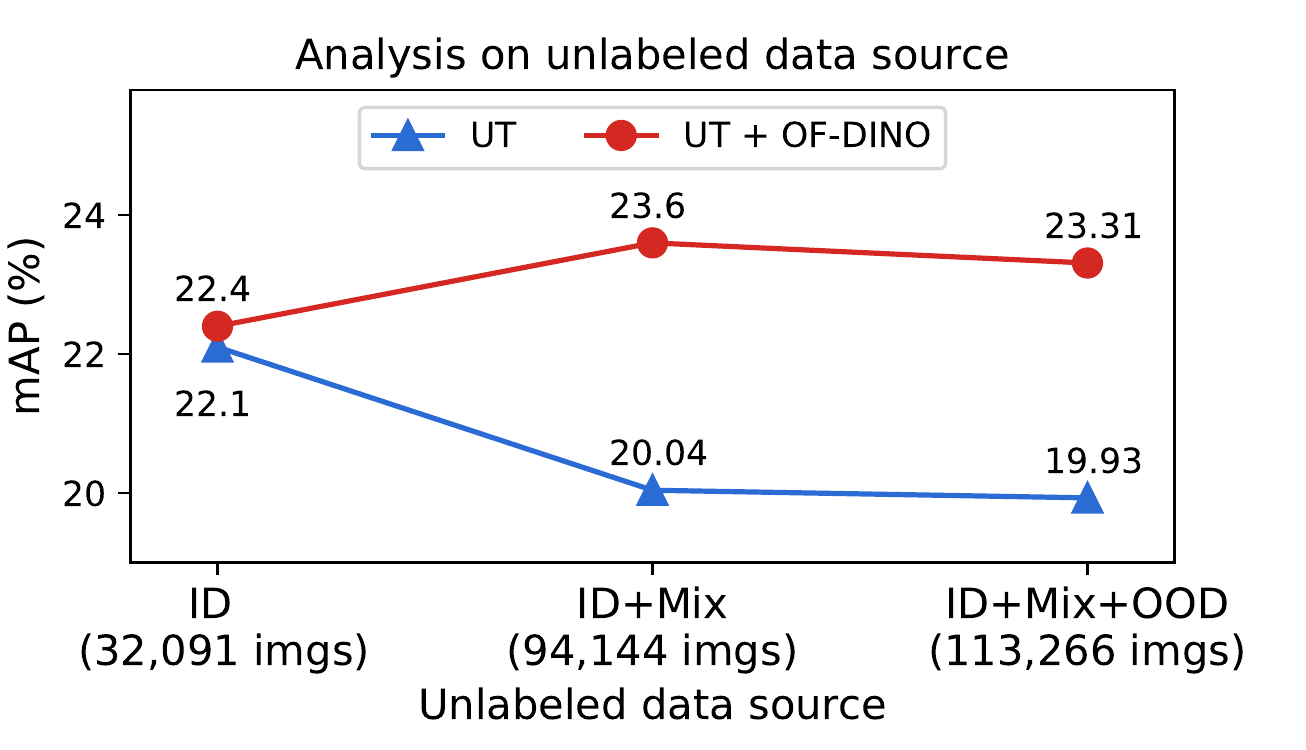}}
\end{minipage}\hfill
\begin{minipage}[t]{0.55\linewidth}
\caption{\small
    Using more unlabeled images containing OOD objects (\textit{i.e.,} Mix and OOD) cannot lead to a better result for UT, while applying our OOD filtering can alleviate the performance drop and improve the accuracy. 
        We experiment on COCO-open with $4k$ images from $40$ ID classes.
}
    \label{fig:datasource}
\end{minipage}
\end{center}
\end{figure}

%% file: table/coco-open.tex
\begin{table}[t]
\centering
\renewcommand{\arraystretch}{1.3}
\caption{
Mean average precision of COCO-Open when \textbf{varying the number of ID objects}. 
We experiment on COCO-Open with $4k$ labeled images from $20/40/60$ ID classes.  We run each method $3$ times and report the standard deviation.}
\label{table:coco-open-cls}
\resizebox{\linewidth}{!}{%
\begin{tabular}{cccc}
\toprule
Num. of ID/OOD objects    & 20/60                  & 40/40                  & 60/20                  \\  \specialrule{1pt}{1pt}{1pt}
Label-only               & 16.89$\pm$2.6          & 15.98$\pm$0.49          & 16.64$\pm$0.59          \\ 
UT         & 18.37$\pm$1.67 \textcolor{myblue}{(+1.48)} & 20.28$\pm$0.85 \textcolor{myblue}{(+4.29)} & 23.09$\pm$0.25 \textcolor{myblue}{(+6.45)} \\
UT + OF-DINO & \textbf{23.43$\pm$2.19 \textcolor{myblue}{(+6.54)}} & \textbf{22.91$\pm$0.28 \textcolor{myblue}{(+6.93)}} & \textbf{24.89$\pm$0.34 \textcolor{myblue}{(+8.25)}} \\ 
\bottomrule
\end{tabular}}
\end{table}

%% file: table/coco-open-num-img.tex
\begin{table}[t]
\centering
\renewcommand{\arraystretch}{1.3}
\caption{Mean average precision of COCO-open under \textbf{different degree of supervision}. 
We experiment on COCO-Open with $1/2/4k$ labeled images from $20$ selected ID classes.  We run each method $3$ times and report the standard deviation.}
\label{table:coco-open-numimg}
\resizebox{\linewidth}{!}{%
\begin{tabular}{cccc}
\toprule
Num. of Labeled Images    & 1,000                  & 2,000                  & 4,000                  \\  \specialrule{1pt}{1pt}{1pt}
Label-only               & 10.20$\pm$ 0.34         & 11.84$\pm$ 0.33         & 16.35$\pm$ 0.28         \\ 
UT         & 11.77$\pm$0.38 \textcolor{myblue}{(+1.57)} & 13.87$\pm$0.68 \textcolor{myblue}{(+2.03)} & 18.23$\pm$0.47 \textcolor{myblue}{(+1.88)} \\
UT + OF-DINO & \textbf{16.80$\pm$0.53 \textcolor{myblue}{(+6.60)}} & \textbf{18.10$\pm$0.71 \textcolor{myblue}{(+6.26)}} & \textbf{22.56$\pm$0.51 \textcolor{myblue}{(+6.21)}} \\\bottomrule
\end{tabular}}
\end{table}

%% file: figure/analysis_fpr.tex
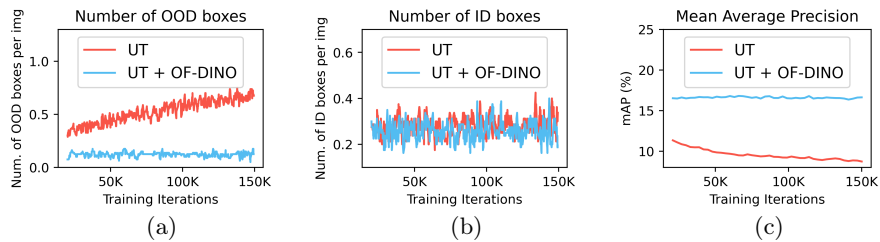
\begin{figure}[t]
   \begin{picture}(0,80)
     \put(5,5){\includegraphics[width=0.3\linewidth]{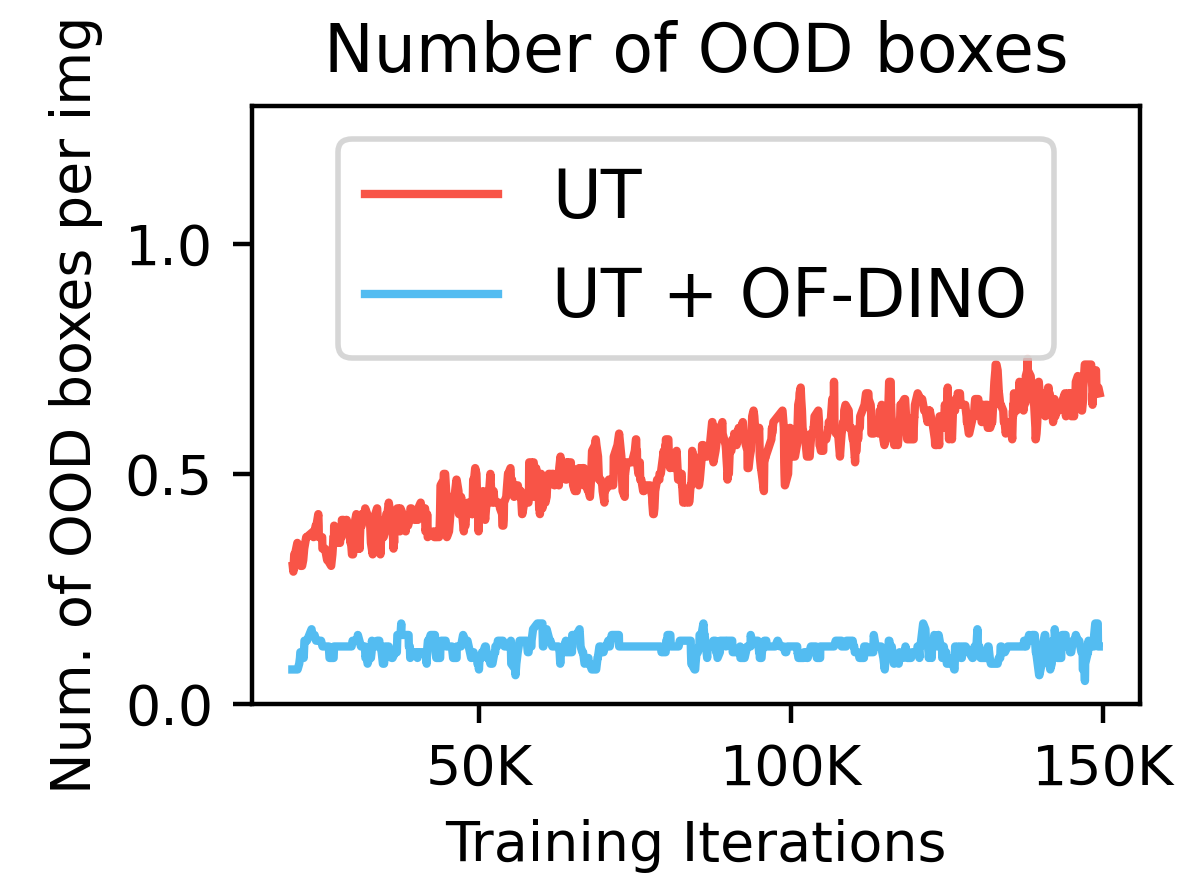}}
     \put(120,5){\includegraphics[width=0.3\linewidth]{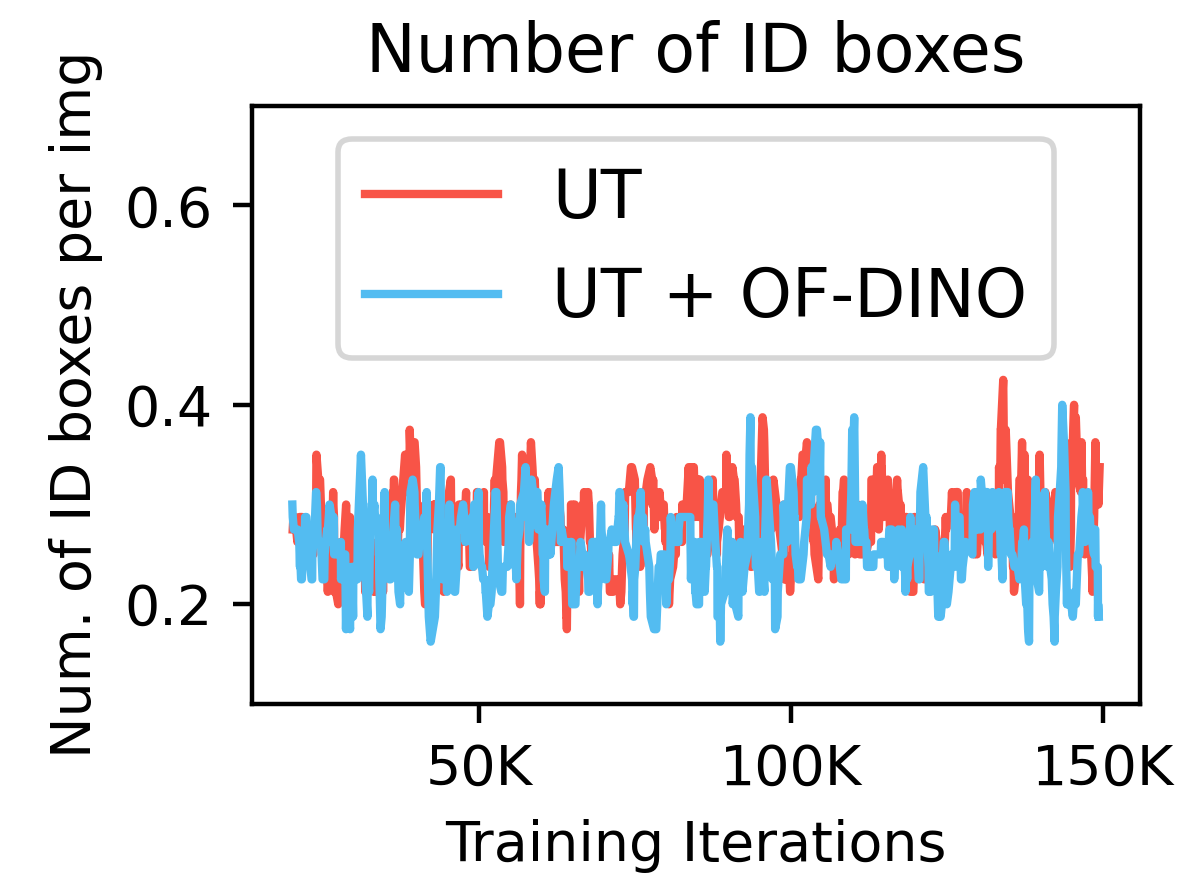}}
     \put(235,5){\includegraphics[width=0.3\linewidth]{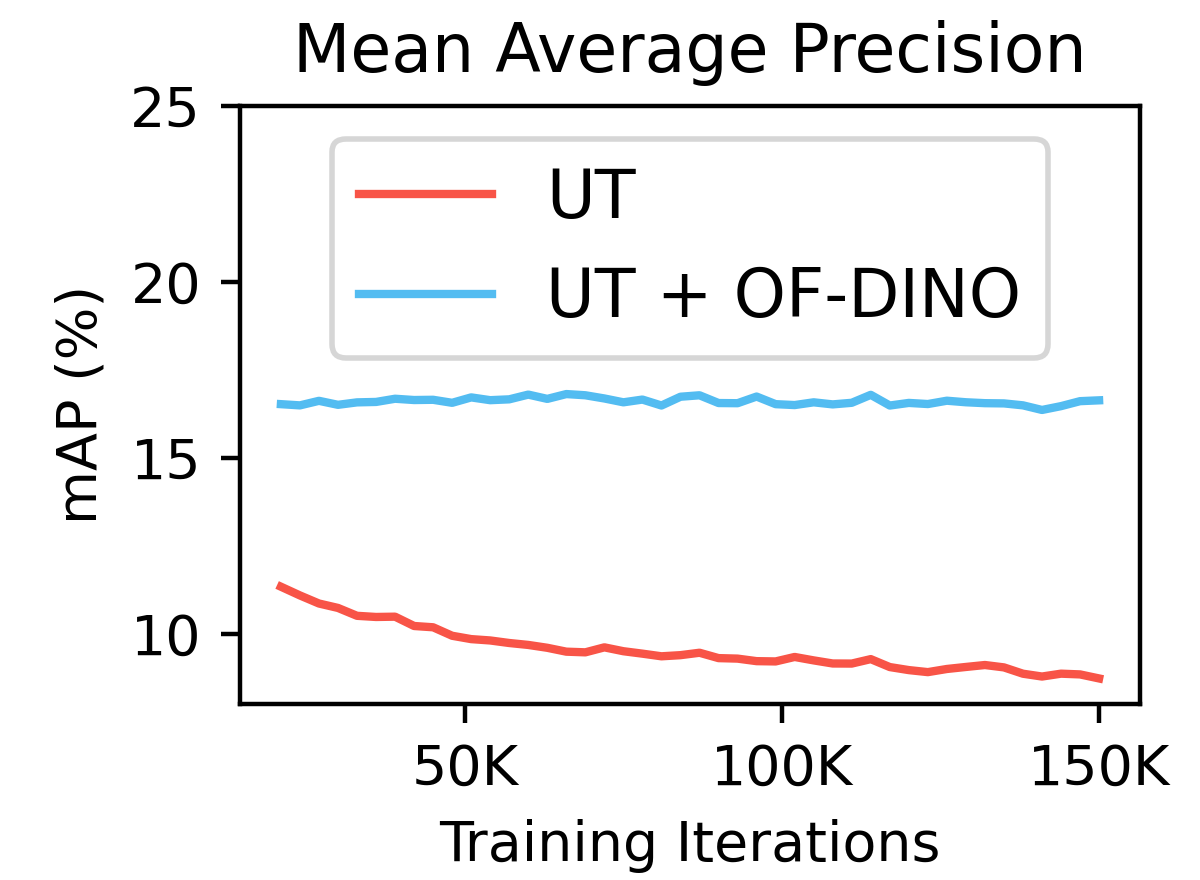}}
     \put(60,-5){(a)}
     \put(175,-5){(b)}
     \put(290,-5){(c)}
   \end{picture}
    \caption{Using OOD detector can (a) suppress the OOD objects while (b) maintain the ID objects in pseudo-labels, and this leads to (c) the improvement on OSSOD. We experiment on COCO-Open with $500$ labeled images from $20$ ID classes.
    }
    \label{fig:analysis_box}
\end{figure}

%% file: table/coco-goi.tex
\begin{table}[t]
\centering
\renewcommand{\arraystretch}{1.3}
\caption{Experimental results of COCO-OpenImages.}
\label{tab:coco_goi}

\resizebox{0.8\linewidth}{!}{

\begin{tabular}{ccccc}
\toprule
 Methods                       & Labeled & Unlabeled   & & mAP   \\ \specialrule{1pt}{1pt}{1pt}
Fully-supervised  & COCO & -                                       &                & 40.90 \\
Fully-supervised  &COCO+OpenImage& -                          &     & 42.91 \\  \midrule
Unbiased Teacher~\cite{liu2021unbiased} &    COCO & OpenImage    &                & 41.81 \\ 
Unbiased Teacher+OF-DINO           &COCO & OpenImage           &            & \textbf{43.14} \\
\bottomrule
\end{tabular}
}
\end{table}

%% file: limitation.tex
\noindent\textbf{Limitations and future works.}
While addressing OSSOD, we do not address other issues such as covariate shift and mismatch in object category distributions between datasets.
The offline OOD detector is an individual module from the object detector, so it requires more computational resources in the training stage.
However, this concern does not exist as we remove the offline OOD detector and only use the object detector in the inference stage. 
Our key message is that combining an offline OOD detection module and an SSOD method is a simple yet effective solution to address OSSOD tasks.
Based on this integrated {framework}, there will be more advanced techniques for both SSOD and OOD detection methods, which can potentially improve the performance on OSSOD tasks.

%% file: conclusion.tex
\section{Conclusion}
Unlike traditional closed-set Semi-Supervised Object Detection, where the majority of the unlabeled images are from in-distribution datasets, we aim to train object detector with \textit{unconstrained} unlabeled images -- Open-Set Semi-Supervised Object Detection (OSSOD).
In our proposed setting, unlabeled training images can contain object classes that are out-of-distribution, \textit{i.e.,} classes not included in the labeled set.
To the best of our knowledge, this is the first work investigating OSSOD.
We present comprehensive studies to understand the challenges for OSSOD and analyze why the performance of existing SoTA methods degrade even when trained with more unlabeled images. 
To overcome the challenges, we consider both online and offline OOD detectors and show that a simple yet effective offline OOD detector can further improve the SSOD models when training with large-scale unlabeled datasets,
\textit{e.g.,} COCO-OpenImage.
We achieved state-of-the-art results across existing and new benchmark settings, and some interesting directions such as more accurate and efficient OOD detectors are worth exploring for future research.

%% file: appedix_section.tex
\appendix
\section{COCO-additional.}
To examine OOD filtering in a large-scale scenario, we consider COCO-additional which aims to improve the fully-supervised object detector with the additional large-scale dataset (\textit{e.g.,} \texttt{COCO2017-unlabeled}).
As presented in Table~\ref{table:cocoadd_sota_complete}, UT (with data augmentation from SoftTeacher~\cite{xu2021end}) can achieve $44.06$ mAP, and using the proposed OOD filtering can further improve UT and achieve $45.14$ mAP and shows state-of-the-art result against the existing SS-OD works~\cite{tang2021humble,zhou2021instant,jeong2019consistency,tang2020proposal,xu2021end} on COCO-additional.
This demonstrates that removing OOD objects from the large-scale unlabeled data can still improve the existing SSOD framework.
However, it is worth noting that our proposed OOD filtering mechanism is not restricted to UT, and we believe that is also complementary to other SSOD methods~\cite{sohn2020simple,tang2021humble,zhou2021instant,jeong2019consistency,tang2020proposal,xu2021end}.

\section{Qualitative Results with the OOD filtering}
\input{figure/qual_result_dino}
To show the effectiveness of the OOD filtering, we show the pseudo-labels generated with and without OOD filtering in Figure~\ref{fig:dino_frcnn_results}.
Without the OOD filtering, some OOD objects (\textit{e.g.,} fish) are predicted as inlier (\textit{e.g.,} bird) with high confidence. 
Such an issue is alleviated by the OOD filtering, which effectively suppresses the OOD objects in pseudo-labels and thus improves the accuracy of the object detector.

\section{Experiment results when using ViT-B as OOD detector}
To understand how the ViT-B pretrained on ImageNet21k performs on OSSOD tasks, we examine the ViT-B model on the experiment setups presented in the main paper. 
Specifically, we consider different degrees of supervision in Table~\ref{table:coco-open-numimg-app}, different numbers of ID objects in Table~\ref{table:coco-open-cls-app}, large-scale SSOD setting (\textit{e.g.,} COCO-additional) in Table~\ref{table:cocoadd_sota_complete}, large-scale OSSOD setting (\textit{e.g.,} COCO-OpenImage) in Table~\ref{table:coco_goi_complete}.
We observe using the ViT-B model can consistently lead to better results in all experiment scenarios, while, as we mentioned in the main paper, ViT-B requires large-scale supervised pre-training dataset (ImageNet21k), which is not suitable for our label-efficient settings. However, it does demonstrate that the method can scale with larger-scale pre-trained models, including when using in-the-wild unlabeled data (e.g. OpenImages). 
\input{table/coco-open-num-img-complete}
\input{table/coco-open-complete}
\input{table/coco-goi-complete}

\section{Experiments on STAC}
\input{table/stac_expts}

In addition to Unbiased Teacher~\cite{liu2021unbiased} experimented with in the main paper, we also consider another SSOD method, STAC~\cite{sohn2020simple}, to show that our findings are general.
As shown in Table~\ref{table:compare_ssod_stac}, we observe STAC also suffers from open-set issues when experimenting on OSSOD tasks. 
Compared with the traditional closed-set SSOD task, the performance gain of STAC is smaller in open-set conditions. 

To address the open-set issues, we also apply our proposed OOD detection method on STAC. 
As presented in Table~\ref{table:ood_dino_stac}, when the OOD detector (DINO) is applied, we can improve STAC from $18.60$ mAP to $19.80$ mAP. 
This shows that our proposed OOD filtering method is not restricted to any particular SSOD method and can potentially improve other SSOD methods. 

Furthermore, similar to Unbiased Teacher~\cite{liu2021unbiased}, other existing SSOD methods~\cite{tang2021humble,zhou2021instant,yang2021interactive,xu2021end} also applied confidence thresholding to select pseudo-labels, so they are also prone to suffer from the semantic expansion issue as we described in the main paper.

\input{figure/analysis-numcls}
\input{figure/coco_goi_stat}

\section{Complete Comparison of OOD Detection}
To compare OOD detection methods, we list a more complete comparison as presented in Table~\ref{table:ood_detection_complete}. 
We list other observations as follows: 
\input{table/coco-open-ood-complete}

\noindent\textbf{Mahalanobis Distance under limited amount of data setting}
With limited amount of training data (\textit{i.e.,} $4$k ID images), Mahalanobis distance becomes unreliable compared with other OOD metrics, and this is contrary to the prior works on OOD detection~\cite{fort2021exploring,lee2018simple}, where a large amount of data is used for training the OOD detectors. 
As computing Mahalanobis distance requires the covariance matrix and the class-wise mean vectors, and estimating the covariance matrix for high-dimension features is difficult and inaccurate, this is even more difficult when the data is scarce (especially when the number of instances is closer to or even less than the number of feature dimensions). 
This is also why we observe that Euclidean distance, which is equivalent to the Mahalanbois distance with the identity convariance matrix, leads to even better results than the Mahalanobis distance. 

Another weakness of Mahalanobis distance is that it requires one to obtain class-wise mean and co-variance features by deriving the features for all ID images in the labeled set, so it is computationally slow (also pointed out in MOS~\cite{huang2021mos}) and thus less preferable for large-scale open-set semi-supervised learning, which requires detecting OOD samples in each training iteration. 

\noindent\textbf{Inverse abstaining confidence under different number of ID/OOD classes.} 
Compared with other offline OOD detection metrics, using inverse abstaining confidence is more robust when varying the number of ID classes. 
To be more specific, as shown in Table~\ref{table:ood_detection_complete}, the Energy score and Shannon entropy perform on par or even better than the inverse abstaining confidence in the case of using $20$ ID classes. 
However, as shown in Figure~\ref{fig:ood_numcls}, when we increase the number of ID classes, the inverse abstaining confidence degrades much less than the Energy score and Shannon entropy. 
Such a property makes the inverse abstaining confidence more suitable for different OSSOD scenarios. 

\section{Comparison between COCO-OpenImage and COCO-additional}
\input{table/minor-cls-goi-coco}

In the main paper, we considered two large-scale unlabeled sets, \texttt{OpenImagev5} and \texttt{COCO2017-unlabeled}, to improve the object detection trained on the labeled \texttt{COCO2017-labeled}.
Our OOD filtering framework improves the supervised object detector from $40.90$ mAP to $43.14$ mAP by using \texttt{OpenImagev5} as an unlabeled set and achieves $45.14$ mAP when the  \texttt{COCO2017-unlabeled} is used as an unlabeled set. 

As \texttt{OpenImagev5} has more unlabeled images than \texttt{COCO2017-unlabeled} ($1.7$M vs. $120$k), we are curious \textit{why the model using the }\texttt{OpenImagev5} \textit{as an unlabeled set cannot outperform the model using the} \texttt{COCO-unlabeled} \textit{as an unlabeled set}.

We attribute this trend to the  following factors:

(i) \textbf{Mismatch in class distribution.} 
We first compare the class distribution of both datasets, and we find these two datasets have very different object distributions as shown in Figure~\ref{fig:coco_goi_dist}\textcolor{red}{a}. 
The mismatch in the class distribution in the unlabeled set is prone to affect the frequency or confidence of objects predicted for the evaluation set, and this potentially leads to performance degradation in the evaluation set.

(ii) \textbf{Some COCO objects are rare in OpenImage.}
When \texttt{OpenImagev5} is used as an unlabeled set to train the object detector in a semi-supervised manner, we observe, as shown in Table~\ref{table:minorcls-coco-goi}, the performance on \textit{some} objects are even lower than the supervised model due to the scarcity of these objects. 
Specifically, even though \texttt{OpenImagev5} has more images than \texttt{COCO2017-train}, the number of some COCO objects in entire \texttt{OpenImage} are even fewer than the objects in \texttt{COCO2017-train}, as shown in Figure~\ref{fig:coco_goi_dist}\textcolor{red}{b}. 
This suggests the objects are very rare and infrequently appear, and such a property potentially limits the further improvement by using \texttt{OpenImagev5}. 
Note that \texttt{COCO2017-unlabeled} follows the same class distribution as \texttt{COCO2017-labeled} (described in COCO official page), and both datasets have similar amount of images ($120$k vs. $117$k).

\section{Label Correspondence between COCO and OpenImage}
\input{table/coco_goi_correspondence}
To construct the baseline trained with ground-truth labels from \texttt{OpenImage}, we manually label the correspondence between $80$ classes in \texttt{MS-COCO} and $601$ classes in \texttt{OpenImage}. 
We provide the object correspondence in Table~\ref{table:coco_goi_matching}.
Among $601$ classes in \texttt{OpenImage}, $139$ GOI classes have matching COCO classes, and the remaining $462$ classes do not correspond to any COCO classes.
We thus remove the labels of these classes in the training of the supervised baseline. 

\section{Implementation Details}
Our implementation is based on the Detectron2 framework. 
As our framework is built on the Unbiased Teacher~\cite{liu2021unbiased}, we follow its implementation details, including training iterations, threshold, unsupervised loss weight, and other hyper-parameters for a fair comparison. 

\textbf{Model Architecture.}
We experiment on the Faster-RCNN with FPN~\cite{lin2017feature}, and ResNet-50 pretrained on ImageNet-1K is used as the feature backbone.
For the offline OOD detectors, we consider DINO~\cite{caron2021emerging} and VITB~\cite{dosovitskiy2020image} as base models. 

\textbf{Training.}
For the object detectors, we use the SGD optimizer with a momentum rate $0.9$ and a learning rate $0.01$, and we use a constant learning rate scheduler for COCO-Open and learning rate decay for COCO-additional and COCO-OpenImage.
Each batch contains $8$ labeled images and $8$ unlabeled images for COCO-Open, and $32$ labeled images and $32$ unlabeled images for COCO-OpenImage and COCO-additional. 
To fine-tune DINO/VITB models, we randomly select $64$ patches from each image and train $10k/20k/40k$ iterations for $1k$/$2k$/$4k$ labeled images setups of COCO-Open.
For COCO-additional and COCO-OpenImage, we also randomly select $64$ patches from each image and train for $160k$ iterations. 
We follow the prior work~\cite{steiner2021train} to use SGD optimizer with a learning rate $1e-3$ and $5e-3$ for DINO and VIT models. 
We apply the inverse abstaining confidence as the OOD score due to its robustness to different number of object categories.
As for thresholds for confidence thresholding and OOD filtering, we use $\delta=0.5$ for the confidence thresholding and $\delta_{ood}=0.5$ for the OOD filtering.

\textbf{Data augmentation.}
For COCO-Open, We follow the data augmentation used in Unbiased Teacher~\cite{liu2021unbiased}, which applies a random horizontal flip for weak augmentation and randomly adds color jittering, grayscale, Gaussian blur, and cutout patches~\cite{devries2017improved} for the strong augmentation.
For COCO-additional and COCO-OpenImage, we additionally consider scale jitter used in SoftTeacher~\cite{xu2021end} to further improve the performance.
Image-level or box-level geometric augmentations, such as rotation, translation, and Mosaic~\cite{zhou2021instant}, are not used in our method.

\section{Training of our online and offline frameworks}
In the main paper, we present our proposed OSSOD framework integrated with online and offline OOD detectors.
We thus present the training details of offline OOD detectors in Alg.~\ref{alg:offline} and online OOD detectors in Alg.~\ref{alg:online}.

\input{train_algorithm}

\input{limitation}

%% file: figure/qual_result_dino.tex
\begin{figure}[h]
   \begin{picture}(0,140)
   \centering
     \put(0,0){\includegraphics[width=\linewidth]{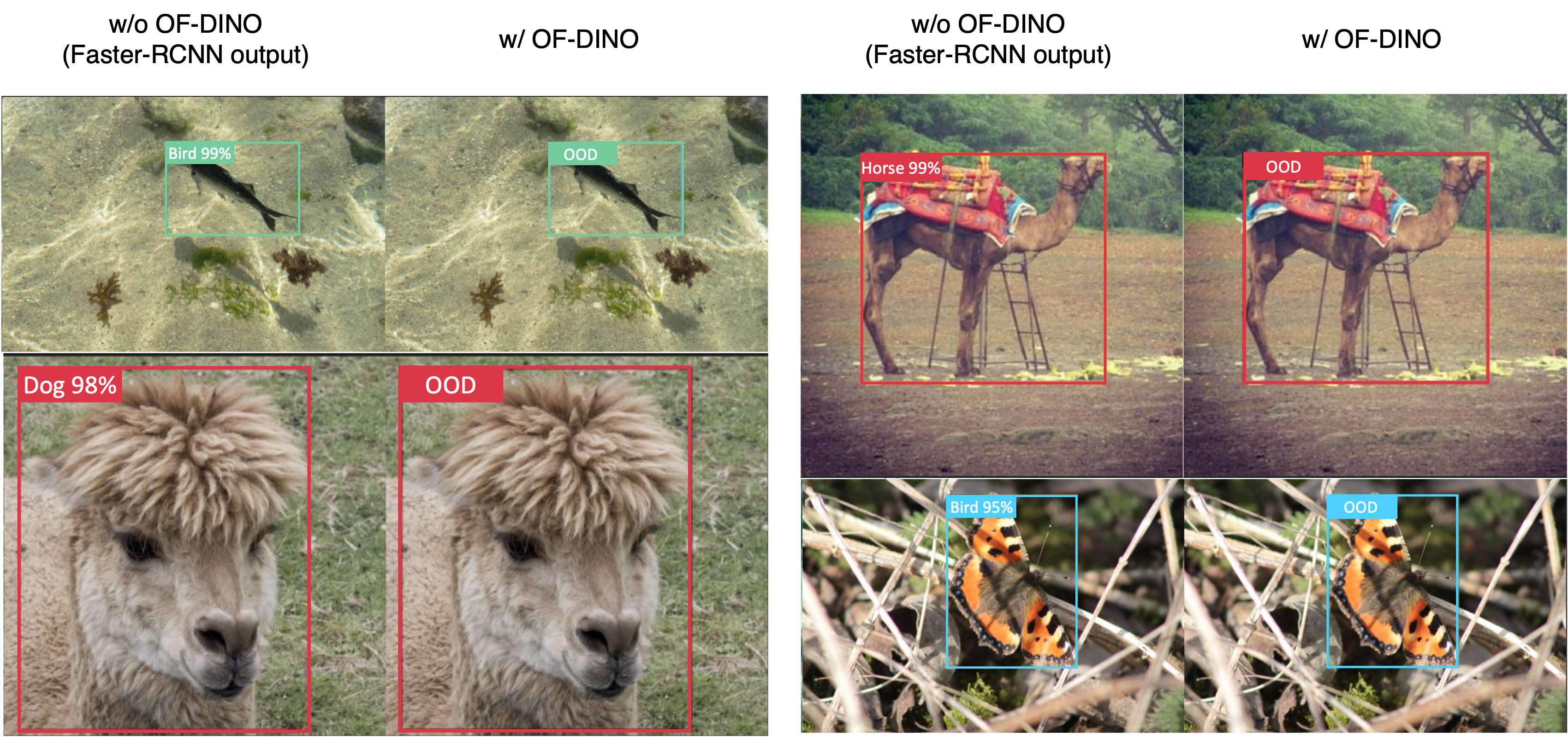}}
   \end{picture}
    \caption{\textbf{Comparison between the pseudo-labels with and without DINO-based OOD filtering (OF-DINO).}}
    \label{fig:dino_frcnn_results}
\end{figure}

%% file: table/coco-open-num-img-complete.tex
\begin{table}[h]
\centering
\renewcommand{\arraystretch}{1.3}
\caption{Mean average precision of COCO-open under \textbf{different degrees of supervision}. We first pre-select $20$ COCO classes as ID classes and $60$ COCO classes as OOD classes, and $1/2/4k$ images are \textit{randomly} selected from the purely-ID set as the labeled set. The remaining images from pure-ID, pure-OOD, and mixed sets are used as the unlabeled set. We run each method $3$ times and report the standard deviation.}
\label{table:coco-open-numimg-app}
\resizebox{0.8\linewidth}{!}{%
\begin{tabular}{clll}
\toprule
Num. of Labeled Images    & 1,000                  & 2,000                  & 4,000                  \\  \specialrule{1pt}{1pt}{1pt}
Label-only               & 10.20$\pm$ 0.34         & 11.84$\pm$ 0.33         & 16.35$\pm$ 0.28         \\ 
UT         & 11.77$\pm$0.38 \textcolor{myblue}{(+1.57)} & 13.87$\pm$0.68 \textcolor{myblue}{(+2.03)} & 18.23$\pm$0.47 \textcolor{myblue}{(+1.88)} \\
UT + OF-DINO & {16.80$\pm$0.53 \textcolor{myblue}{(+6.60)}} & {18.10$\pm$0.71 \textcolor{myblue}{(+6.26)}} & {22.56$\pm$0.51 \textcolor{myblue}{(+6.21)}} \\ \midrule
UT + OF-ViT  & \textbf{17.10$\pm$0.46 \textcolor{myblue}{(+6.90)}} & \textbf{19.32$\pm$0.53 \textcolor{myblue}{(+7.48)}} & \textbf{23.01$\pm$0.67 \textcolor{myblue}{(+6.66)}}\\\bottomrule
\end{tabular}}
\end{table}

%% file: table/coco-open-complete.tex
\begin{table}[h]
\centering
\renewcommand{\arraystretch}{1.3}
\caption{
Mean average precision of COCO-Open when \textbf{varying the number of ID objects}. We first randomly sample $20/40/60$ COCO classes as ID classes and remaining COCO classes as OOD classes, and $4k$ images are randomly selected from the purely-ID set as the labeled set. The remaining images from pure-ID, pure-OOD, and mixed sets are used as the unlabeled set. We run each method $3$ times and report the standard deviation.}
\label{table:coco-open-cls-app}
\resizebox{0.8\linewidth}{!}{%
\begin{tabular}{clll}
\toprule
Num. of ID/OOD objects    & 20/60                  & 40/40                  & 60/20                  \\  \specialrule{1pt}{1pt}{1pt}
Label-only               & 16.89$\pm$2.6          & 15.98$\pm$0.49          & 16.64$\pm$0.59          \\ 
UT         & 18.37$\pm$1.67 \textcolor{myblue}{(+1.48)} & 20.28$\pm$0.85 \textcolor{myblue}{(+4.29)} & 23.09$\pm$0.25 \textcolor{myblue}{(+6.45)} \\
UT + OF-DINO & {23.43$\pm$2.19 \textcolor{myblue}{(+6.54)}} & {22.91$\pm$0.28 \textcolor{myblue}{(+6.93)}} & {24.89$\pm$0.34 \textcolor{myblue}{(+8.25)}} \\ \midrule
UT + OF-ViT  & \textbf{25.20$\pm$2.00 \textcolor{myblue}{(+8.31)}} & \textbf{25.10$\pm$1.01 \textcolor{myblue}{(+9.12)}} & \textbf{26.11$\pm$0.40 \textcolor{myblue}{(+9.47)}}\\
\bottomrule
\end{tabular}}
\end{table}

%% file: table/coco-goi-complete.tex
\begin{table}[h]
\begin{minipage}{.45\linewidth}
\centering
\renewcommand{\arraystretch}{1.2}
\caption{
Comparison to other SSOD methods in COCO-additional.
}

\resizebox{0.75\linewidth}{!}{
\begin{tabular}{cc}
\toprule
                                              & mAP   \\ 
                                              \midrule
Supervised                                                     & 40.90 \\ \midrule
Proposal Learning~\cite{tang2020proposal}                      & 38.40 \\
CSD~\cite{jeong2019consistency}                                & 38.82 \\
STAC~\cite{sohn2020simple}                                     & 39.21 \\
Instant-Teaching~\cite{zhou2021instant}                        & 40.20 \\
MOCOv2 + Instagram-1B~\cite{tang2021humble}                    & 41.10 \\
Humble Teacher~\cite{tang2021humble}                           & 42.37 \\
SoftTeacher~\cite{xu2021end}                                   & 44.05 \\
Unbiased Teacher*~\cite{liu2021unbiased}                       & 44.06 \\ 
Unbiased Teacher* + OF-DINO                                    & {45.14} \\ \midrule
Unbiased Teacher* + OF-ViT                                   & \textbf{45.16} \\ 
\bottomrule
\end{tabular}
}
\label{table:cocoadd_sota_complete}
\end{minipage}
\begin{minipage}{.02\linewidth}
\centering
\renewcommand{\arraystretch}{1.2}
\resizebox{1.0\linewidth}{!}{
\begin{tabular}{@{\extracolsep{4pt}}c@{}}
\toprule
\bottomrule
\end{tabular}
}
\end{minipage}
\begin{minipage}{.5\linewidth}
\centering
\renewcommand{\arraystretch}{1.2}
\caption{Experimental results of COCO-OpenImage.}
\resizebox{1.0\linewidth}{!}{

\begin{tabular}{ccc}
\toprule
                                        & OpenImage GT labels    & mAP   \\ \specialrule{1pt}{1pt}{1pt}
COCO                                    &                       & 40.90 \\
COCO + OpenImage                        & \checkmark            & 42.91 \\  \midrule
Unbiased Teacher~\cite{liu2021unbiased} &                       & 41.81 \\ 
Unbiased Teacher + OF-DINO           &                       & {43.14} \\ \midrule
Unbiased Teacher + OF-ViT           &                       & \textbf{43.48} \\ 
\bottomrule
\end{tabular}

}
\label{table:coco_goi_complete}
\end{minipage}
\end{table}

%% file: table/stac_expts.tex
\begin{table}[h]
\centering
\caption{
\textbf{Generalization of our findings to other methods, namely STAC~\cite{sohn2020simple} performance comparison between closed-set SSOD and open-set SSOD.} For closed-set SSOD, we randomly select $1\%/2\%$ from the training set (\textit{i.e.,} $1172/2234$ labeled images). For the open-set SSOD, we randomly samples 20 classes as ID classes and the remaining classes as OOD classes; hence the differences in performance of ``Labeled only''. We then sample the same amount of labeled images in both cases for a fair comparison.
}
\label{table:compare_ssod_stac}
\resizebox{0.5\linewidth}{!}{%
\begin{tabular}{ccccccc}
\toprule
                 & \multicolumn{2}{c}{closed-set SSOD}& & \multicolumn{2}{c}{open-set SSOD}                           \\\cmidrule{2-3} \cmidrule{5-6}
Percentage of  labeled images           & 1$\%$       & 2$\%$              & & 1$\%$          & 2$\%$             \\ 
Num. of labeled images                  & 1,172       & 2,344              & & 1,172           & 2,344              \\ \specialrule{1pt}{1pt}{1pt}
Labeled only                            &  9.05       &    12.70           & &  11.20          &   12.18                  \\
STAC~\cite{sohn2020simple}                &  13.97      & 18.25            & & 13.22           &    15.34              \\ \midrule
$\Delta$                                &  \textcolor{myblue}{+4.92}       &   \textcolor{myblue}{+5.55}     &    & \textcolor{myblue}{+2.02 }&    \textcolor{myblue}{+3.16 }      \\ 
\bottomrule
\end{tabular}
}
\end{table}
\begin{table}[h]
\centering
\renewcommand{\arraystretch}{1.3}
\caption{\textbf{OOD filtering improves STAC on COCO-Open.} We randomly sample $40$ COCO classes as ID classes and remaining COCO classes as OOD classes, and $4k$ images are randomly selected from the purely-ID set as the labeled set. The remaining images from pure-ID, pure-OOD, and mixed sets are used as the unlabeled set.}
\label{table:ood_dino_stac}
\resizebox{0.5\linewidth}{!}{%
\begin{tabular}{cccc}
\toprule
    & Label-only      & STAC       & STAC + OF-DINO                  \\  \specialrule{1pt}{1pt}{1pt}
mAP                       & 16.54           &  18.60 \textcolor{myblue}{(+2.06)}     & 19.80 \textcolor{myblue}{(+3.26)}          \\ 
\bottomrule
\end{tabular}}
\end{table}

%% file: figure/analysis-numcls.tex
\begin{figure}[ht]
   \begin{picture}(0,110)
     \put(60,5){\includegraphics[width=0.3\linewidth]{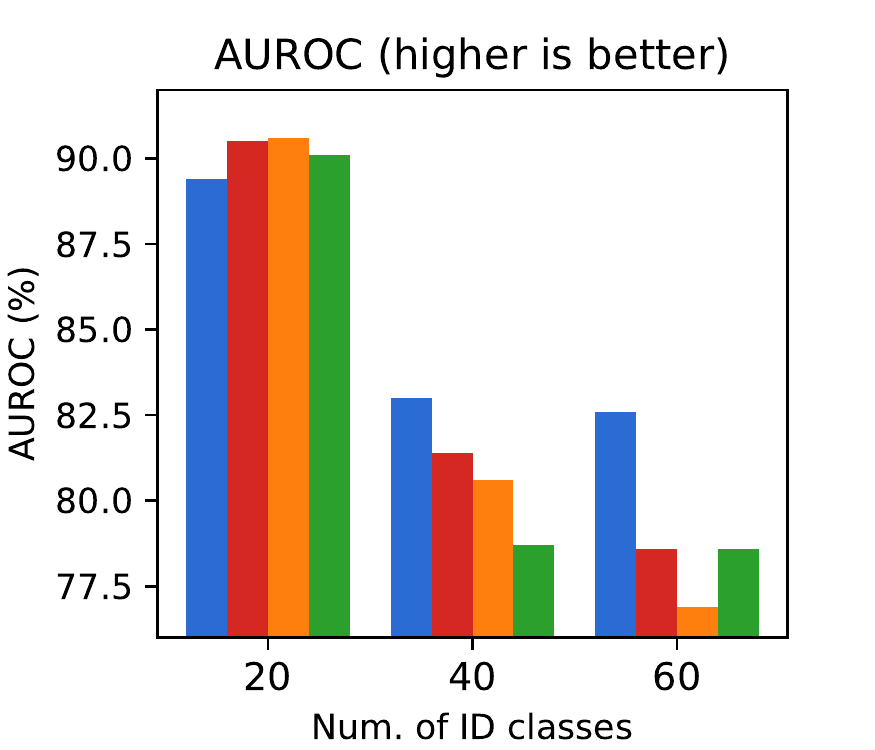}}
     \put(200,5){\includegraphics[width=0.3\linewidth]{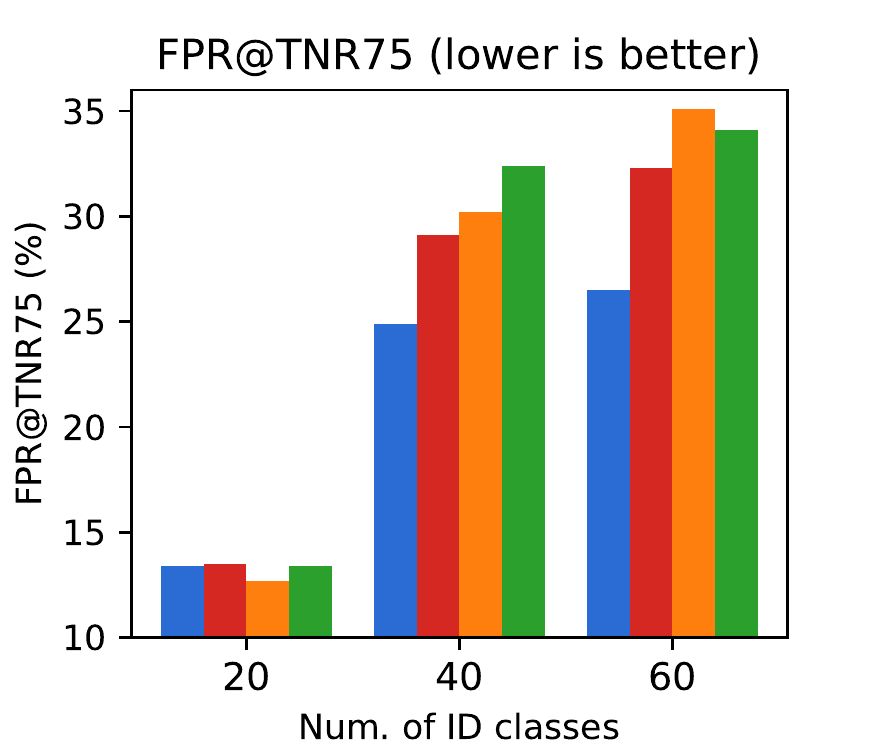}}
     \put(80,95){\includegraphics[width=0.6\linewidth]{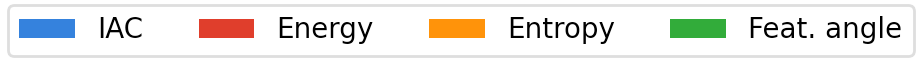}}
     \put(110,-3){(a)}
     \put(250,-3){(b)}
   \end{picture}
    \caption{Comparison of OOD metrics of OF-DINO under different number of ID classes, and we present (a) AUROC and (b) FPR@TNR75 to evaluate the performance of the OOD detection. Among different OOD metrics, inverse abstaining confidence (\textcolor{myblue}{IAC}) suffers less when the number of ID classes increases.  Note that $20/40/60$ classes from the \texttt{COCO2017-train} are selected as ID classes (and the remaining classes as OOD classes), and $4k$ images of pure-ID set are selected as the labeled set.}
    \label{fig:ood_numcls}
\end{figure}

%% file: figure/coco_goi_stat.tex
\begin{figure}[h]
   \begin{picture}(0,280)
   \centering
     \put(-10,150){\includegraphics[width=1\linewidth]{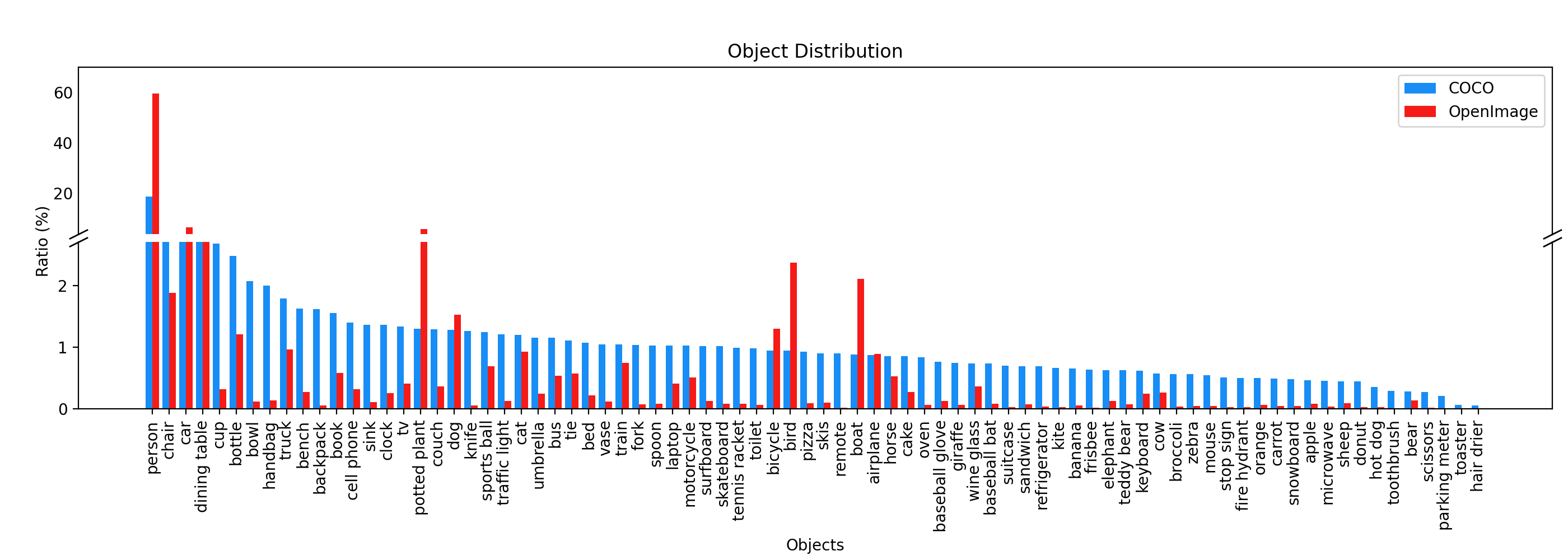}}
     \put(-10,10){\includegraphics[width=1\linewidth]{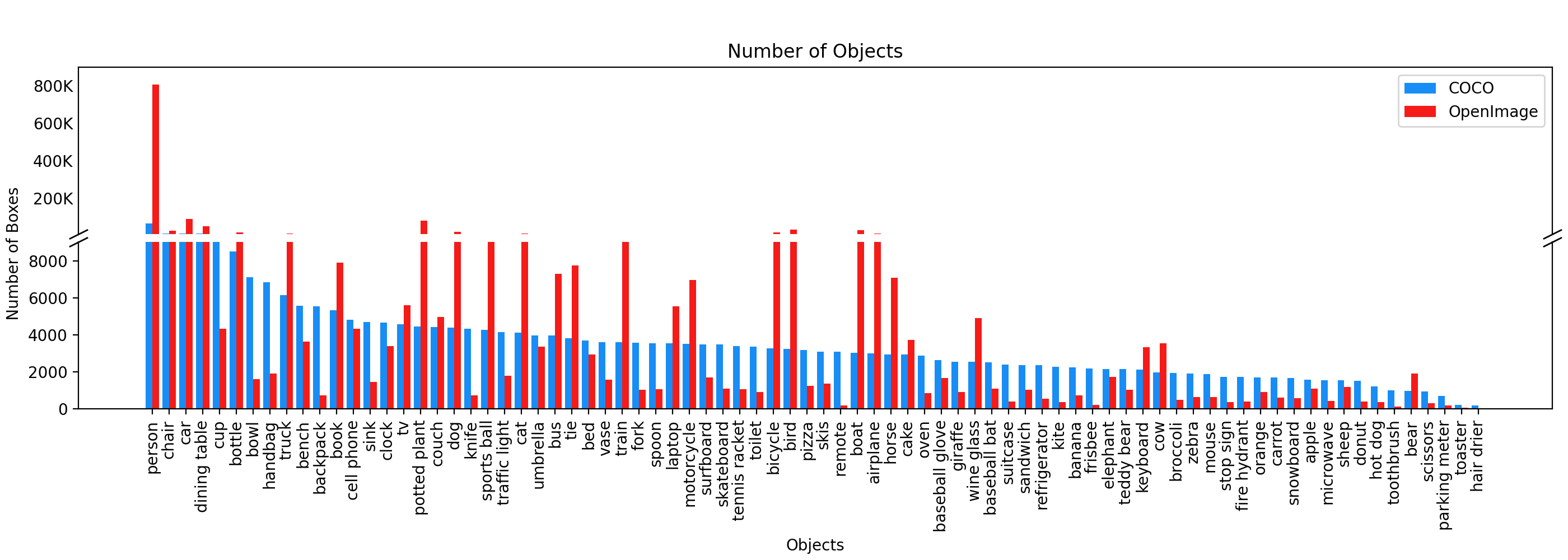}}
     \put(165,0){(b)}
     \put(165,140){(a)}
   \end{picture}
    \caption{(a) Distribution of object categories and (b) number of objects  in \texttt{COCO2017-train} and \texttt{OpenImagev5}.}
    \label{fig:coco_goi_dist}
\end{figure}

%% file: table/coco-open-ood-complete.tex
\begin{table*}[ht]
\centering
\renewcommand{\arraystretch}{1.3}
\caption{Evaluation of OOD detection for object detection tasks. We sample $20$ classes from COCO as in-distribution (ID) objects, and  $4000$ pure-ID images are selected as labeled images. All methods are evaluated on \texttt{COCO2017-val}. Value before the slashes indicates ignoring background patches when computing AUROC and FPR, and value after the slashes regradining background patches as OOD objects when computing AUROC and FPR.}

\label{table:ood_detection_complete}
\resizebox{\linewidth}{!}{%
\begin{tabular}{ccccccc}
\toprule
Model                        & Methods                    & OoD Scores $\gamma_{ood}$    & AUROC $\uparrow$ & FPR50$\downarrow$ & FPR75$\downarrow$ & FPR95$\downarrow$ \\

\midrule
\multirow{9}{*}{\begin{tabular}[c]{@{}c@{}}Online OOD Detector \\ (Faster-RCNN branch)\end{tabular}} & \multirow{5}{*}{Vanilla}               & MSP~\cite{hendrycks2017baseline}   & 67.0        / 71.0        & 22.5        / 15.5        & 58.4        / 52.4        & 92.3        / 91.1        \\
                                                  &                                        & Energy~\cite{liu2021energy}      & 75.5        / 68.2        & 13.2        / 22.8        & 36.8        / 49.0        & 83.6        / 87.8        \\
                                                  &                                        & Entropy     & 75.9        / 68.4        & 12.2        / 22.3        & 38.5        / 51.1        & 83.1        / 87.7        \\
                                                  &                                        & Mahalanobis~\cite{lee2018simple} & 50.2        / 61.6        & 51.5        / 32.8        & 83.0        / 65.7        & 98.1        / 93.7        \\
                                                  &                                        & Euclidean   & 56.3        / 61.5        & 40.2        / 31.8        & 74.3        / 66.9        & 96.1        / 94.1        \\
                                                  & OE~\cite{hendrycks2019oe}& MSP         & 67.0        / 73.3        & 25.3        / 15.2        & 55.0        / 45.9        & 89.1        / 85.6        \\
                                                  & One-vs-all~\cite{saito2021openmatch}   & MSP         & 73.0        / 76.0        & 13.4        / 10.7        & 45.7        / 40.2        & 90.0        / 84.8        \\
                                                  & GODIN~\cite{hsu2020generalized}        & Cosine $h(x)$  & 77.8        / 73.5        & 12.5        / 14.6        & 33.8        / 45.0       & 77.4        / 84.5        \\
                                                  & GSD~\cite{tian2021geometric}           & Feat. angle & 78.7        / 71.3        & 11.8        / 19.3        & 32.1        / 48.8        & 73.9        / 83.4        \\ \midrule
\multirow{5}{*}{\begin{tabular}[c]{@{}c@{}}Offline OOD Detector\\ (DINO)\end{tabular}}                             & \multirow{5}{*}{Ours}  & Inv. abstaining conf.     & 83.6        / 86.0        & 8.5         / \textbf{5.9}         & 22.4        / 18.7        & 61.7        / 57.9        \\
                                                  &                                        & Energy      & 89.6        / 85.9        & 4.0         / 7.0         & 12.2        / 18.8        & 47.5        / 56.8        \\
                                                  &                                        & Entropy     & 88.9        / 84.7        & \textbf{3.5}         / 7.3         & 12.6        / 20.3        & 51.1        / 59.9        \\
                                                  &                                        & Mahalanobis~\cite{lee2018simple} & 81.8        / 75.7        & 11.7        / 17.6        & 25.6        / 35.9        & 57.6        / 68.9        \\
                                                  &                                        & Euclidean   & \textbf{90.8}        / \textbf{86.1}        & 3.6        / 7.3         & \textbf{10.7}        / \textbf{18.5}        & \textbf{38.6}        / \textbf{51.6}        \\ \midrule
\multirow{4}{*}{\begin{tabular}[c]{@{}c@{}}Offline OOD Detector\\ (ViT)\end{tabular}}                               & \multirow{4}{*}{Ours}  & Inv. abstaining conf.         & 87.5        / 88.2        & 4.5         / 4.0         & 15.3        / 14.7        & 54.7        / 51.5        \\
                                                  &                                        & Energy      & 93.3        / 88.5        & 2.1         / 5.9         & 6.0         / 14.5        & 32.4        / 45.3        \\
                                                  &                                        & Entropy     & 93.2        / 88.1        & 1.5         / 5.8         & 5.9         / 15.1        & 33.9        / 46.3        \\
                                                  &                                        & Feat. angle       & 93.2        / 87.6        & 2.1         / 7.2         & 6.1         / 15.5        & 33.4        / 46.0.      \\
              \bottomrule
\end{tabular}
}
\end{table*}

%% file: table/minor-cls-goi-coco.tex
\begin{table*}[h]
\centering
\renewcommand{\arraystretch}{1.3}
\caption{\textbf{Performance degradation of minor objects in semi-supervised learning.} We apply Unbiased Teacher with the proposed OOD filtering (DINO) and use the \texttt{OpenImagev5} as an unlabeled set, and detection performance of some rare objects are degraded due to the scarcity of these objects in \texttt{OpenImagev5}. Note that \texttt{OpenImagev5} contains $1.7$M images, and \texttt{COCO2017-train} has $117$k images.  }
\label{table:minorcls-coco-goi}
\resizebox{\linewidth}{!}{%
\begin{tabular}{cccc|cc}
\toprule
Objects       & \begin{tabular}[c]{@{}c@{}}\textbf{Supervised-only}\\ Labeled: \texttt{COCO2017-train}\\ - \end{tabular} & \begin{tabular}[c]{@{}c@{}}\textbf{UT+OF-DINO}\\ Labeled: \texttt{COCO2017-train}\\ Unlabeled: \texttt{OpenImagev5}\end{tabular} & mAP difference & \begin{tabular}[c]{@{}c@{}}Number of boxes in\\ \texttt{COCO2017-train}\end{tabular} & \begin{tabular}[c]{@{}c@{}}Number of boxes in\\ \texttt{OpenImagev5}\end{tabular} \\\midrule
sheep         & 51.81                                                                             & 51.49                                                                                              & \textcolor{red}{-0.33}          & 1529                                                                        & 1188                                                                     \\
carrot        & 22.52                                                                             & 22.34                                                                                              & \textcolor{red}{-0.18}          & 1683                                                                        & 594                                                                      \\
hair drier    & 2.59                                                                              & 1.53                                                                                               & \textcolor{red}{-1.06}         & 189                                                                         & 27                                                                       \\
zebra         & 66.63                                                                             & 65.99                                                                                              & \textcolor{red}{-0.63}          & 1916                                                                        & 621                                                                      \\
snowboard     & 35.48                                                                             & 33.90                                                                                              & \textcolor{red}{-1.57}          & 1654                                                                        & 574                                                                      \\
knife         & 19.60                                                                             & 19.45                                                                                              & \textcolor{red}{-0.15}          & 4326                                                                        & 726                                                                      \\
banana        & 23.56                                                                             & 23.08                                                                                              & \textcolor{red}{-0.48}          & 2243                                                                        & 723                                                                      \\
orange        & 32.29                                                                             & 31.38                                                                                              & \textcolor{red}{-0.91}          & 1699                                                                        & 900                                                                      \\
hot dog       & 32.23                                                                             & 31.20                                                                                              & \textcolor{red}{-1.03}          & 1222                                                                        & 362                                                                      \\
toaster       & 40.40                                                                             & 34.28                                                                                              & \textcolor{red}{-6.13}          & 217                                                                         & 60                                                                       \\
giraffe       & 68.41                                                                             & 68.04                                                                                              & \textcolor{red}{-0.37}          & 2546                                                                        & 920                                                                      \\
tennis racket & 49.31                                                                             & 49.10                                                                                              & \textcolor{red}{-0.22}          & 3394                                                                        & 1047                                                                     \\
microwave     & 54.14                                                                             & 53.75                                                                                              & \textcolor{red}{-0.40}          & 1547                                                                        & 432                                                                      \\\bottomrule
\end{tabular}
}
\end{table*}

%% file: table/coco_goi_correspondence.tex
\begin{table}[h]
\centering
\renewcommand{\arraystretch}{1.3}
\caption{Object classes correspondence between \texttt{MS-COCO} and \texttt{OpenImages}. $139$ OpenImages objects have the matching COCO objects, while the remaining $462$ OpenImage objects do not correspond to any COCO object.}
\label{table:coco_goi_matching}
\resizebox{0.9\linewidth}{!}{%
\begin{tabular}{clcl}
\toprule
\textbf{COCO-objects}   & \textbf{OpenImage-objects}                                                                                                        & \textbf{COCO-objects}   & \textbf{OpenImage-objects}                                            \\\midrule
person         & Person, Boy, Woman, Man, Girl                                                                                     & wine glass   & Wine glass                                           \\
bicycle        & Bicycle                                                                                                           & cup          & Coffee cup, Measuring cup, Mug                       \\
car            & Car, Ambulance, Limousine, Taxi                                                                                   & fork         & Fork                                                 \\
motorcycle     & Motorcycle                                                                                                        & knife        & Knife, Kitchen knife                                 \\
airplane       & Airplane                                                                                                          & spoon        & Spoon, Ladle                                         \\
bus            & Bus                                                                                                               & bowl         & Mixing bowl, Bowl                                    \\
train          & Train                                                                                                             & banana       & Banana                                               \\
truck          & Truck, Van                                                                                                        & apple        & Apple                                                \\
boat           & Boat, Barge, Gondola, Canoe                                                                                       & sandwich     & Submarine sandwich, Sandwich                         \\
traffic light  & Traffic light                                                                                                     & orange       & Orange                                               \\
fire hydrant   & Fire hydrant                                                                                                      & broccoli     & Broccoli                                             \\
stop sign      & Stop sign                                                                                                         & carrot       & Carrot                                               \\
parking meter  & Parking meter                                                                                                     & hot dog      & Hot dog                                              \\
bench          & Bench                                                                                                             & pizza        & Pizza                                                \\
cat            & Cat                                                                                                               & cake         & Cake                                                 \\
dog            & Dog                                                                                                               & chair        & Chair                                                \\
horse          & Horse                                                                                                             & couch        & Studio couch, Couch, Sofa bed, Loveseat              \\
sheep          & Sheep                                                                                                             & potted plant & Lavender (Plant), Plant, Houseplant, Flowerpot       \\
cow            & Cattle, Bull                                                                                                      & bed          & Bed                                                  \\
elephant       & Elephant                                                                                                          & dining table & Kitchen \& dining room table, Table, Coffee table    \\
bear           & Bear, Brown bear, Panda, Polar bear                                                                               & toilet       & Toilet, Bidet                                        \\
zebra          & Zebra                                                                                                             & tv           & Computer monitor, Television                         \\
giraffe        & Giraffe                                                                                                           & laptop       & Laptop                                               \\
backpack       & Backpack                                                                                                          & mouse        & Computer mouse                                       \\
umbrella       & Umbrella                                                                                                          & remote       & Remote control                                       \\
handbag        & Handbag                                                                                                           & keyboard     & Computer keyboard                                    \\
tie            & Tie                                                                                                               & cell phone   & Mobile phone                                         \\
suitcase       & Suitcase                                                                                                          & microwave    & Microwave oven                                       \\
frisbee        & Flying disc                                                                                                       & oven         & Oven, Gas stove                                      \\
skis           & Ski                                                                                                               & toaster      & Toaster                                              \\
snowboard      & Snowboard                                                                                                         & sink         & Sink                                                 \\
donut         & Doughnut                                                                                                           & refrigerator & Refrigerator                                         \\
kite           & Kite                                                                                                              & book         & Book                                                 \\
baseball bat   & Baseball bat                                                                                                      & clock        & Wall clock, Clock, Alarm clock, Digital clock, Watch \\
baseball glove & Baseball glove                                                                                                    & vase         & Vase                                                 \\
skateboard     & Skateboard                                                                                                        & scissors     & Scissors                                             \\
surfboard      & Surfboard                                                                                                         & teddy bear   & Teddy bear                                           \\
tennis racket  & Tennis racket, Racket                                                                                             & hair drier   & Hair dryer                                           \\
bottle         & Beer, Bottle, Wine                                                                                                & toothbrush   & Toothbrush                                          \\
\multirow{2}{*}{\begin{tabular}[c]{@{}c@{}}bird\end{tabular}}           & Bird, Magpie, Woodpecker, Blue jay, Raven, Eagle,         & \multirow{2}{*}{\begin{tabular}[c]{@{}c@{}}sports ball\end{tabular}}   & Rugby ball, Football, Ball, Cricket ball, \\
               & Falcon, Owl, Duck, Canary, Goose, Swan, Parrot, Sparrow         &         & Volleyball (Ball), Golf ball, Tennis ball                                             \\
\bottomrule
\end{tabular}
}
\end{table}

%% file: train_algorithm.tex
\begin{algorithm}[H]
\small
\KwData{Labeled set: $D_{s}=\{x_s, y_s\}$; Unlabeled set: $D_{u}=\{x^u\}$
}
  \For{ Iters. of supervised training an object detector}{
    Compute supervised object detector loss $\mathcal{L}_{sup}$ with $D_s$\\
    $\theta^{s}_{obj} \leftarrow \theta^{s}_{obj} - \nabla_{\theta^{s}_{obj}} \mathcal{L}_{sup}$
  }
    $\theta^{t}_{obj} \leftarrow \theta^{s}_{obj}$\\
  \For{ Iters. of training an offline OOD detector}{
    Get background proposal boxes $\tilde{y}_t$ from Teacher object detector\\
    Select foreground GT boxes from $y^s$ and crop $I_{fg}$ from $x_s$\\
    Select background proposal boxes from $\tilde{y}_t$ and crop $I_{bg}$ from $x_s$\\
    Compute multi-class cross-entropy loss $\mathcal{L}_{ood}$ based on \{$I_{bg}, I_{fg}$\}\\
    $\theta_{ood} \leftarrow \theta_{ood} - \nabla_{\theta_{ood}} \mathcal{L}_{ood}$
  }
  \For{Iters. of semi-supervised training of an object detector}{
    Predict $\tilde{y}_u = f(x_u; \theta_t)$ \\
    Apply confidence thresholding $\hat{y}_u \leftarrow h(\tilde{y}_u; \delta)$ \\
    Apply OOD filtering $\bar{y}_u \leftarrow h(\hat{y}_u; \delta_{ood})$ \\
    Compute unsupervised object detector loss $\mathcal{L}_{unsup}$ with $\{D_u, \bar{y}_u \}$\\
    Compute supervised object detector loss $\mathcal{L}_{sup}$ with $D_s$\\
    $\mathcal{L}_{ssod} = \mathcal{L}_{sup}+\lambda\mathcal{L}_{unsup}$\\
    $\theta^{s}_{obj} \leftarrow \theta^{s}_{obj} - \nabla_{\theta^{s}_{obj}} \mathcal{L}_{ssod}$\\
    $\theta^{t}_{obj} \leftarrow \alpha \theta^{t}_{obj} + (1-\alpha) \theta^{s}_{obj}$\\
  }
\KwResult{Learned weights $\theta^{t}_{obj}$/$\theta^{s}_{obj}$ of Teacher/Student object detector}
\caption{Learning of an \textbf{offline OOD Detector} and UT~\cite{liu2021unbiased}}\label{alg:offline}
\normalsize
\end{algorithm}

\begin{algorithm}[H]
\small
\KwData{Labeled set: $D_{s}=\{x_s, y_s\}$; Unlabeled set: $D_{u}=\{x^u\}$
}
  Add an OOD detection head on both Teacher and Student object detectors\\
  \For{Iters. of semi-supervised training of an object detector}{
    Predict $\tilde{y}_u = f(x_u; \theta_t)$ \\
    Apply confidence thresholding $\hat{y}_u \leftarrow h(\tilde{y}_u; \delta)$ \\
    Apply OOD filtering $\bar{y}_u \leftarrow h(\hat{y}_u; \delta_{ood})$ \\
    Compute unsupervised object detector loss $\mathcal{L}_{unsup}$ with $\{D_u, \bar{y}_u \}$\\
    Compute supervised object detector loss $\mathcal{L}_{sup}$ with $D_s$\\
    Compute OOD loss $\mathcal{L}_{ood}$ (Refer to definition in original papers)\\
    $\mathcal{L} = \mathcal{L}_{sup}+\lambda\mathcal{L}_{unsup} + \lambda_{ood}\mathcal{L}_{ood}$\\
    $\theta^{s}_{obj} \leftarrow \theta^{s}_{obj} - \nabla_{\theta^{s}_{obj}} \mathcal{L}$\\
    $\theta^{t}_{obj} \leftarrow \alpha \theta^{t}_{obj} + (1-\alpha) \theta^{s}_{obj}$\\
  }
\KwResult{Learned weights $\theta^{t}_{obj}$/$\theta^{s}_{obj}$ of Teacher/Student object detector}
\caption{Learning of an \textbf{online OOD Detector} and UT~\cite{liu2021unbiased}}\label{alg:online}
\normalsize
\end{algorithm}

%% file: main.bbl
\begin{thebibliography}{10}
\providecommand{\url}[1]{\texttt{#1}}
\providecommand{\urlprefix}{URL }
\providecommand{\doi}[1]{https://doi.org/#1}

\bibitem{berthelot2019mixmatch}
Berthelot, D., Carlini, N., Goodfellow, I., Papernot, N., Oliver, A., Raffel,
  C.A.: Mixmatch: A holistic approach to semi-supervised learning. In: Advances
  in Neural Information Processing Systems (NeurIPS). pp. 5049--5059 (2019)

\bibitem{caron2021emerging}
Caron, M., Touvron, H., Misra, I., J{\'e}gou, H., Mairal, J., Bojanowski, P.,
  Joulin, A.: Emerging properties in self-supervised vision transformers. arXiv
  preprint arXiv:2104.14294  (2021)

\bibitem{mmdetection}
Chen, K., Wang, J., Pang, J., Cao, Y., Xiong, Y., Li, X., Sun, S., Feng, W.,
  Liu, Z., Xu, J., Zhang, Z., Cheng, D., Zhu, C., Cheng, T., Zhao, Q., Li, B.,
  Lu, X., Zhu, R., Wu, Y., Dai, J., Wang, J., Shi, J., Ouyang, W., Loy, C.C.,
  Lin, D.: {MMDetection}: Open mmlab detection toolbox and benchmark. arXiv
  preprint arXiv:1906.07155  (2019)

\bibitem{devries2017improved}
DeVries, T., Taylor, G.W.: Improved regularization of convolutional neural
  networks with cutout. arXiv preprint arXiv:1708.04552  (2017)

\bibitem{dhamija2020overlooked}
Dhamija, A., Gunther, M., Ventura, J., Boult, T.: The overlooked elephant of
  object detection: Open set. In: Proceedings of the IEEE Winter Conference on
  Applications of Computer Vision (WACV) (2020)

\bibitem{dosovitskiy2020image}
Dosovitskiy, A., Beyer, L., Kolesnikov, A., Weissenborn, D., Zhai, X.,
  Unterthiner, T., Dehghani, M., Minderer, M., Heigold, G., Gelly, S., et~al.:
  An image is worth 16x16 words: Transformers for image recognition at scale.
  In: Proceedings of the International Conference on Learning Representations
  (ICLR) (2021)

\bibitem{du2022vos}
Du, X., Wang, Z., Cai, M., Li, Y.: Vos: Learning what you don't know by virtual
  outlier synthesis. arXiv preprint arXiv:2202.01197  (2022)

\bibitem{fort2021exploring}
Fort, S., Ren, J., Lakshminarayanan, B.: Exploring the limits of
  out-of-distribution detection. In: Advances in Neural Information Processing
  Systems (NeurIPS) (2021)

\bibitem{girish2021towards}
Girish, S., Suri, S., Rambhatla, S.S., Shrivastava, A.: Towards discovery and
  attribution of open-world gan generated images. In: Proceedings of the
  IEEE/CVF International Conference on Computer Vision. pp. 14094--14103 (2021)

\bibitem{gu2021open}
Gu, X., Lin, T.Y., Kuo, W., Cui, Y.: Open-vocabulary object detection via
  vision and language knowledge distillation. In: Proceedings of the
  International Conference on Learning Representations (ICLR) (2022)

\bibitem{guo2019mixup}
Guo, H., Mao, Y., Zhang, R.: Mixup as locally linear out-of-manifold
  regularization. In: Proceedings of the AAAI Conference on Artificial
  Intelligence (AAAI). vol.~33, pp. 3714--3722 (2019)

\bibitem{hendrycks2017baseline}
Hendrycks, D., Gimpel, K.: A baseline for detecting misclassified and
  out-of-distribution examples in neural networks. In: Proceedings of the
  International Conference on Learning Representations (ICLR) (2017)

\bibitem{hendrycks2019oe}
Hendrycks, D., Mazeika, M., Dietterich, T.: Deep anomaly detection with outlier
  exposure. In: Proceedings of the International Conference on Learning
  Representations (ICLR) (2019)

\bibitem{hendrycks2020augmix}
Hendrycks, D., Mu, N., Cubuk, E.D., Zoph, B., Gilmer, J., Lakshminarayanan, B.:
  {AugMix}: A simple data processing method to improve robustness and
  uncertainty. Proceedings of the International Conference on Learning
  Representations (ICLR)  (2020)

\bibitem{hsu2020generalized}
Hsu, Y.C., Shen, Y., Jin, H., Kira, Z.: Generalized odin: Detecting
  out-of-distribution image without learning from out-of-distribution data. In:
  Proceedings of the IEEE Conference on Computer Vision and Pattern Recognition
  (CVPR) (2020)

\bibitem{huang2021trash}
Huang, J., Fang, C., Chen, W., Chai, Z., Wei, X., Wei, P., Lin, L., Li, G.:
  Trash to treasure: Harvesting ood data with cross-modal matching for open-set
  semi-supervised learning. In: Proceedings of the IEEE/CVF International
  Conference on Computer Vision. pp. 8310--8319 (2021)

\bibitem{huang2021mos}
Huang, R., Li, Y.: Mos: Towards scaling out-of-distribution detection for large
  semantic space. In: Proceedings of the IEEE Conference on Computer Vision and
  Pattern Recognition (CVPR) (2021)

\bibitem{huynh2021open}
Huynh, D., Kuen, J., Lin, Z., Gu, J., Elhamifar, E.: Open-vocabulary instance
  segmentation via robust cross-modal pseudo-labeling. arXiv preprint
  arXiv:2111.12698  (2021)

\bibitem{jeong2019consistency}
Jeong, J., Lee, S., Kim, J., Kwak, N.: Consistency-based semi-supervised
  learning for object detection. In: Advances in Neural Information Processing
  Systems (NeurIPS) (2019)

\bibitem{joseph2021towards}
Joseph, K., Khan, S., Khan, F.S., Balasubramanian, V.N.: Towards open world
  object detection. In: Proceedings of the IEEE Conference on Computer Vision
  and Pattern Recognition (CVPR) (2021)

\bibitem{kim2022learning}
Kim, D., Lin, T.Y., Angelova, A., Kweon, I.S., Kuo, W.: Learning open-world
  object proposals without learning to classify. IEEE Robotics and Automation
  Letters  (2022)

\bibitem{OpenImages2}
Krasin, I., Duerig, T., Alldrin, N., Ferrari, V., Abu-El-Haija, S., Kuznetsova,
  A., Rom, H., Uijlings, J., Popov, S., Kamali, S., Malloci, M., Pont-Tuset,
  J., Veit, A., Belongie, S., Gomes, V., Gupta, A., Sun, C., Chechik, G., Cai,
  D., Feng, Z., Narayanan, D., Murphy, K.: Openimages: A public dataset for
  large-scale multi-label and multi-class image classification. Dataset
  available from https://storage.googleapis.com/openimages/web/index.html
  (2017)

\bibitem{laine2016temporal}
Laine, S., Aila, T.: Temporal ensembling for semi-supervised learning. In:
  Proceedings of the International Conference on Learning Representations
  (ICLR) (2017)

\bibitem{lee2018simple}
Lee, K., Lee, K., Lee, H., Shin, J.: A simple unified framework for detecting
  out-of-distribution samples and adversarial attacks. In: Advances in Neural
  Information Processing Systems (NeurIPS) (2018)

\bibitem{liang2018odin}
Liang, S., Li, Y., Srikant, R.: Enhancing the reliability of
  out-of-distribution image detection in neural networks. In: Proceedings of
  the International Conference on Learning Representations (ICLR) (2018)

\bibitem{lin2017feature}
Lin, T.Y., Doll{\'a}r, P., Girshick, R., He, K., Hariharan, B., Belongie, S.:
  Feature pyramid networks for object detection. In: Proceedings of the IEEE
  Conference on Computer Vision and Pattern Recognition (CVPR) (2017)

\bibitem{lin2014microsoft}
Lin, T.Y., Maire, M., Belongie, S., Hays, J., Perona, P., Ramanan, D.,
  Doll{\'a}r, P., Zitnick, C.L.: Microsoft coco: Common objects in context. In:
  Proceedings of the European Conference on Computer Vision (ECCV) (2014)

\bibitem{liu2021energy}
Liu, W., Wang, X., Owens, J.D., Li, Y.: Energy-based out-of-distribution
  detection. In: Advances in Neural Information Processing Systems (NeurIPS)
  (2020)

\bibitem{liu2021unbiased}
Liu, Y.C., Ma, C.Y., He, Z., Kuo, C.W., Chen, K., Zhang, P., Wu, B., Kira, Z.,
  Vajda, P.: Unbiased teacher for semi-supervised object detection. In:
  Proceedings of the International Conference on Learning Representations
  (ICLR) (2021)

\bibitem{luo2021consistency}
Luo, H., Cheng, H., Gao, Y., Li, K., Zhang, M., Meng, F., Guo, X., Huang, F.,
  Sun, X.: On the consistency training for open-set semi-supervised learning.
  arXiv preprint arXiv:2101.08237  (2021)

\bibitem{miller2021uncertainty}
Miller, D., S{\"u}nderhauf, N., Milford, M., Dayoub, F.: Uncertainty for
  identifying open-set errors in visual object detection. arXiv preprint
  arXiv:2104.01328  (2021)

\bibitem{mohseni2020self}
Mohseni, S., Pitale, M., Yadawa, J., Wang, Z.: Self-supervised learning for
  generalizable out-of-distribution detection. In: Proceedings of the AAAI
  Conference on Artificial Intelligence (AAAI) (2020)

\bibitem{nalisnick2018deep}
Nalisnick, E., Matsukawa, A., Teh, Y.W., Gorur, D., Lakshminarayanan, B.: Do
  deep generative models know what they don't know? In: Proceedings of the
  International Conference on Learning Representations (ICLR) (2019)

\bibitem{pidhorskyi2018generative}
Pidhorskyi, S., Almohsen, R., Adjeroh, D.A., Doretto, G.: Generative
  probabilistic novelty detection with adversarial autoencoders. In: Advances
  in Neural Information Processing Systems (NeurIPS) (2018)

\bibitem{ren2015faster}
Ren, S., He, K., Girshick, R., Sun, J.: Faster r-cnn: Towards real-time object
  detection with region proposal networks. In: Advances in neural information
  processing systems (NeurIPS). pp. 91--99 (2015)

\bibitem{sabokrou2018adversarially}
Sabokrou, M., Khalooei, M., Fathy, M., Adeli, E.: Adversarially learned
  one-class classifier for novelty detection. In: Proceedings of the IEEE
  Conference on Computer Vision and Pattern Recognition (CVPR) (2018)

\bibitem{saito2021learning}
Saito, K., Hu, P., Darrell, T., Saenko, K.: Learning to detect every thing in
  an open world. arXiv preprint arXiv:2112.01698  (2021)

\bibitem{saito2021openmatch}
Saito, K., Kim, D., Saenko, K.: Openmatch: Open-set consistency regularization
  for semi-supervised learning with outliers. In: Advances in Neural
  Information Processing Systems (NeurIPS) (2021)

\bibitem{sajjadi2016regularization}
Sajjadi, M., Javanmardi, M., Tasdizen, T.: Regularization with stochastic
  transformations and perturbations for deep semi-supervised learning. In:
  Advances in Neural Information Processing Systems (NeurIPS). pp. 1163--1171
  (2016)

\bibitem{sohn2020fixmatch}
Sohn, K., Berthelot, D., Li, C.L., Zhang, Z., Carlini, N., Cubuk, E.D.,
  Kurakin, A., Zhang, H., Raffel, C.: Fixmatch: Simplifying semi-supervised
  learning with consistency and confidence. In: Advances in Neural Information
  Processing Systems (NeurIPS) (2020)

\bibitem{sohn2020simple}
Sohn, K., Zhang, Z., Li, C.L., Zhang, H., Lee, C.Y., Pfister, T.: A simple
  semi-supervised learning framework for object detection. arXiv preprint
  arXiv:2005.04757  (2020)

\bibitem{steiner2021train}
Steiner, A., Kolesnikov, A., Zhai, X., Wightman, R., Uszkoreit, J., Beyer, L.:
  How to train your vit? data, augmentation, and regularization in vision
  transformers. arXiv preprint arXiv:2106.10270  (2021)

\bibitem{tang2020proposal}
Tang, P., Ramaiah, C., Xu, R., Xiong, C.: Proposal learning for semi-supervised
  object detection. arXiv preprint arXiv:2001.05086  (2020)

\bibitem{tang2021humble}
Tang, Y., Chen, W., Luo, Y., Zhang, Y.: Humble teachers teach better students
  for semi-supervised object detection. In: Proceedings of the IEEE/CVF
  Conference on Computer Vision and Pattern Recognition. pp. 3132--3141 (2021)

\bibitem{tarvainen2017mean}
Tarvainen, A., Valpola, H.: Mean teachers are better role models:
  Weight-averaged consistency targets improve semi-supervised deep learning
  results. In: Advances in neural information processing systems (NeurIPS). pp.
  1195--1204 (2017)

\bibitem{thulasidasan2021effective}
Thulasidasan, S., Thapa, S., Dhaubhadel, S., Chennupati, G., Bhattacharya, T.,
  Bilmes, J.: An effective baseline for robustness to distributional shift.
  arXiv preprint arXiv:2105.07107  (2021)

\bibitem{tian2021geometric}
Tian, J., Yung, D., Hsu, Y.C., Kira, Z.: A geometric perspective towards neural
  calibration via sensitivity decomposition. In: Advances in Neural Information
  Processing Systems (NeurIPS) (2021)

\bibitem{wu2019detectron2}
Wu, Y., Kirillov, A., Massa, F., Lo, W.Y., Girshick, R.: Detectron2.
  \url{https://github.com/facebookresearch/detectron2} (2019)

\bibitem{xu2021end}
Xu, M., Zhang, Z., Hu, H., Wang, J., Wang, L., Wei, F., Bai, X., Liu, Z.:
  End-to-end semi-supervised object detection with soft teacher. arXiv preprint
  arXiv:2106.09018  (2021)

\bibitem{yang2021interactive}
Yang, Q., Wei, X., Wang, B., Hua, X.S., Zhang, L.: Interactive self-training
  with mean teachers for semi-supervised object detection. In: Proceedings of
  the IEEE Conference on Computer Vision and Pattern Recognition (CVPR) (2021)

\bibitem{yu2020multi}
Yu, Q., Ikami, D., Irie, G., Aizawa, K.: Multi-task curriculum framework for
  open-set semi-supervised learning. In: Proceedings of the European Conference
  on Computer Vision (ECCV) (2020)

\bibitem{yun2019cutmix}
Yun, S., Han, D., Oh, S.J., Chun, S., Choe, J., Yoo, Y.: Cutmix: Regularization
  strategy to train strong classifiers with localizable features. In:
  Proceedings of the IEEE International Conference on Computer Vision (ICCV).
  pp. 6023--6032 (2019)

\bibitem{zareian2021open}
Zareian, A., Rosa, K.D., Hu, D.H., Chang, S.F.: Open-vocabulary object
  detection using captions. In: Proceedings of the IEEE Conference on Computer
  Vision and Pattern Recognition (CVPR) (2021)

\bibitem{zhang2018mixup}
Zhang, H., Cisse, M., Dauphin, Y.N., Lopez-Paz, D.: mixup: Beyond empirical
  risk minimization. In: Proc. International Conference on Learning
  Representations (ICLR) (2018)

\bibitem{zhou2021instant}
Zhou, Q., Yu, C., Wang, Z., Qian, Q., Li, H.: Instant-teaching: An end-to-end
  semi-supervised object detection framework. In: Proceedings of the IEEE
  Conference on Computer Vision and Pattern Recognition (CVPR) (2021)

\bibitem{zhou2022detecting}
Zhou, X., Girdhar, R., Joulin, A., Kr{\"a}henb{\"u}hl, P., Misra, I.: Detecting
  twenty-thousand classes using image-level supervision. arXiv preprint
  arXiv:2201.02605  (2022)

\bibitem{zhu2019soft}
Zhu, C., Chen, F., Shen, Z., Savvides, M.: Soft anchor-point object detection.
  In: Proceedings of the European Conference on Computer Vision (ECCV) (2020)

\bibitem{zong2018deep}
Zong, B., Song, Q., Min, M.R., Cheng, W., Lumezanu, C., Cho, D., Chen, H.: Deep
  autoencoding gaussian mixture model for unsupervised anomaly detection. In:
  Proceedings of the International Conference on Learning Representations
  (ICLR) (2018)

\end{thebibliography}
